\documentclass[journal]{IEEEtran}
\usepackage{amsmath,amssymb,amsfonts}
\usepackage{algorithmic}
\usepackage{textcomp}
\usepackage{xcolor}
\usepackage{booktabs}
\usepackage{siunitx}
\usepackage{threeparttable}
\usepackage{hyperref}
\usepackage{cleveref}
\usepackage{enumitem}
\usepackage{booktabs}
\usepackage{multirow}
\usepackage{rotating}
\usepackage[normalem]{ulem}
\usepackage[utf8]{inputenc}

\newcommand{\etal}{\emph{et al.}}

\usepackage{ifpdf}
\usepackage{cite}
\ifCLASSINFOpdf
  \DeclareGraphicsExtensions{.pdf,.jpeg,.png}
\else
\fi
\usepackage{amsmath}
\ifCLASSOPTIONcompsoc
 \usepackage[caption=false,font=normalsize,labelfont=sf,textfont=sf]{subfig}
\else
 \usepackage[caption=false,font=footnotesize]{subfig}
\fi
\usepackage{url}

\begin{document}

\title{Vau da Muntanialas: Energy-Efficient Multi-Die Scalable Acceleration of RNN Inference}

\author{Gianna~Paulin,~\IEEEmembership{Student Member,~IEEE,}
        Francesco~Conti,~\IEEEmembership{Member,~IEEE,}
        Lukas~Cavigelli,~\IEEEmembership{Member,~IEEE,}
        and~Luca~Benini,~\IEEEmembership{Fellow,~IEEE}%
\thanks{This work was supported in part by the Heterogenous Computing Systems with Customized Accelerators Project by the Swiss National Science Foundation as part of the Croatian-Swiss Research Program under Grant IZHRZ0\_180625}%
\thanks{Gianna Paulin is with the Integrated System Laboratory, ETH Z\"urich, 8092 Z\"urich, Switzerland (e-mail: pauling@iis.ee.ethz.ch)}%
\thanks{Francesco Conti is with the Department of Electrical, Electronic and Information Engineering, University of Bologna, 40136 Bologna, Italy (e-mail: f.conti@unibo.it)}%
\thanks{Lukas Cavigelli is with Zurich Research Center, Huawei Technologies, 8050 Z\"urich, Switzerland (e-mail: lukas.cavigelli@huawei.com).}%
\thanks{Luca Benini is with the Integrated System Laboratory, ETH Z\"urich, 8092 Z\"urich, Switzerland, and also with the Department of Electrical, Electronic and Information Engineering, University of Bologna, 40136 Bologna, Italy (e-mail: luca.benini@unibo.it).}%
\thanks{Manuscript accepted July 14, 2021; published on July 30, 2021 by IEEE Transactions on Circuits and Systems---I: Regular Papers}%
\thanks{Peer-reviewed and published article is available at \url{https://doi.org/10.1109/TCSI.2021.3099716}.}%
\thanks{Digital Object Identifier 10.1109/TCSI.2021.3099716}}

\markboth{DOI 10.1109/TCSI.2021.3099716}%
{Paulin \MakeLowercase{\textit{et al.}}: Vau Da Muntanialas}

\IEEEpubid{1549-8328~\copyright~2021 IEEE. Personal use is permitted, but republication/redistribution requires IEEE permission.}

\maketitle

\begin{abstract}
Recurrent neural networks such as Long Short-Term Memories (LSTMs) learn temporal dependencies by keeping an internal state, making them ideal for time-series problems such as speech recognition.
However, the output-to-input feedback creates distinctive memory bandwidth and scalability challenges in designing accelerators for RNNs.
We present \textsc{Muntaniala}, an RNN accelerator architecture for LSTM inference with a silicon-measured energy-efficiency of 3.25\,TOP/s/W and performance of 30.53\,GOP/s in UMC 65nm technology.
The scalable design of \textsc{Muntaniala} allows running large RNN models by combining multiple tiles in a systolic array.
We keep all parameters stationary on every die in the array, drastically reducing the I/O communication to only loading new features and sharing partial results with other dies.
For quantifying the overall system power, including I/O power, we built \textsc{Vau da Muntanialas}, to the best of our knowledge, the first demonstration of a systolic multi-chip-on-PCB array of RNN accelerator.
Our multi-die prototype performs LSTM inference with 192 hidden states in 330\si{\micro\second} with a total system power of 9.0\si{\milli\watt} at 10\si{\mega\hertz} consuming 2.95\si{\micro\joule}.
Targeting the 8/16-bit quantization implemented in \textsc{Muntaniala}, we show a phoneme error rate (PER) drop of approximately 3\% with respect to floating-point (FP) on a 3L-384NH-123NI LSTM network on the TIMIT dataset.
\end{abstract}


\IEEEpeerreviewmaketitle

%%%%%%%%%%%%%%%%%%%%%%%%%%%%%%%%%%%%%%%%%%%%%%%%%%%%%%%%%%%%%%%%%%%%%%%%%%%%%%%%%%%%%%%%
%%%%%%%%%%%%%%%%%%%%%%%%%%%%%%%%%%%%%%%%%%%%%%%%%%%%%%%%%%%%%%%%%%%%%%%%%%%%%%%%%%%%%%%%
\section{Introduction}

\IEEEPARstart{T}{he} availability of vast amounts of training data and computing power has enabled increasingly sophisticated machine learning (ML) algorithms, particularly models based on deep learning (DL), to excel in many tasks, such as image recognition~\cite{He2016}, speech recognition~\cite{Hannun2014}, natural language processing~\cite{Young2018}, language translation~\cite{Vaswani2017}, and autonomous gaming bots~\cite{Silver2017}.
While a lot of DL research focuses on image processing, one particular time series problem has seen much interest in both industry and research: automatic speech recognition (ASR).
The emerge of Deep Neural Network (DNN) for ASR has enabled novel speech-based user interfaces such as Amazon Alexa, Google Assistant, Apple Siri, Microsoft Cortana, and others.
Recurrent neural networks (RNNs) are DL models that include an internal state allowing them to learn time-dependencies, making them an ideal candidate for learning tasks on time series, like ASR.
Even though new time series oriented models based on, e.g., convolutional neural networks (CNNs)~\cite{Bai2018} or attention (e.g., Transformer~\cite{Vaswani2017}) have been proposed, RNNs, and especially LSTMs, were still 21\% of all DL training workload of Google's tensor processing units (TPUs) in their datacenters in 2019~\cite{Jouppi2020}.
Especially time series applications targeting medium to small Internet-of-Things (IoT) devices with limited memory and compute resources have been mainly addressed by RNNs.
In these scenarios, a part of the ASR alongside many other tasks (e.g., Keyword Spotting) is still primarily approached using two particular RNN types: Long Short-Term Memory (LSTM) and Gated Recurrent Units (GRUs).

\IEEEpubidadjcol
The leading-edge accuracy of these methods has motivated a strong push toward embedded low-power accelerators to bring these advantages to energy-constrained products such as hearing aids or headphones~\cite{Chen2015}.
As research has paid a lot of attention to CNNs, many specialized FPGA\cite{wang2017efficient,Yuan2021} and ASIC~\cite{scherer2020cutie,lin2017data,Conti2018a,Chen2016,Andri2017} accelerators for low-power CNN inference have been proposed, which achieve energy-efficiency gains in the range of three orders of magnitude with respect to general-purpose architectures~\cite{Reuther2019}.
A significant contributor to this significant energy-efficiency gain comes from computing not in floating-point but rather in fixed-point numbers, which makes specialized training methods for minimizing an imposed accuracy loss an absolute requirement.
While recent research shows that CNNs are very resilient towards accuracy loss even when using binary weights, there is less work available on quantizing RNNs, whose training process is less stable and more complex than for CNNs~\cite{Salehinejad2017}.
Additionally, the specialized architectural optimizations developed for accelerating CNN inference cannot directly be adopted to accelerate RNN inference. 
The additional challenges, such as the necessity of storing and regularly updating an internal state, the densely connected layers with a low computation to memory-footprint ratio, ask for novel algorithmic and architectural solutions.
State-of-the-art LSTM RNN models for ASR can contain up to multiple millions of parameters~\cite{Salehinejad2017}, making efficient data transfer and storage design a necessity.
Typically, there are two possibilities: 
all parameters and internal states are either stored on-chip (e.g., SRAMs), which results in large dies dominated by SRAM, or stored off-chip in dense DRAM memories, whose content needs to be constantly (re-)loaded into the accelerator.
The former approach of scaling the accelerator by arbitrarily increasing the die size is not cost-effective and ultimately infeasible due to decreasing yield.
The second option implies slower and energy-inefficient state data access. Hence, performance and energy become dominated by the constant, high I/O activity toward external memory.

In this work, we present a solution based on the on-chip storage approach that allows scaling beyond a single die to keep the data local, readily accessible in a more energy-efficient way than with the constant weight reloading approach for bigger problem sizes.
Building upon our recently presented energy-efficient LSTM accelerator called \textsc{Chipmunk}~\cite{Conti2018}, this work presents the following contributions:

\begin{enumerate}
    \item We introduce \textsc{Muntaniala}\footnote{Romansh for ``marmot''.}: an extension of \textsc{Chipmunk} for easier integration in a systolic array.
    The architecture of the \textsc{Muntaniala} chip-tile allows multiple tiles to work together in a multi-chip n$\times$n array, while keeping all network parameters on-die local, and minimizing inter-die traffic, resulting in an ideal solution for the cost limited die scaling.
    \item We present \textsc{Vau da Muntanialas}\footnote{Romansh for ``marmot burrow''.}, to the best of our knowledge, the first full-system demonstration of an exemplary systolic grid of $2\times 2$ \textsc{Muntaniala} LSTM accelerator chips on an FPGA-controlled PCB, collaboratively performing LSTM inference with $192$ hidden states (using tiny dies capable of storing only $96$ hidden states each) with minimal total system power of 9.0 \si{\milli\watt} over 330\si{\micro\second} at 10MHz, 1.2V core supply and 2.5V pad supply.
    \item In contrast to most publications on accelerators that ignore I/O power, we measured the I/O and core power consumption of our prototypes, allowing us to make a complete power evaluation.
    \item We have trained an LSTM network on the TIMIT dataset and studied the quantization losses for the chosen 8/16-bit quantization used for \textsc{Muntaniala}.
\end{enumerate}

The next section \Cref{sec:relatedwork} gives an overview of the related work.
The following architecture \Cref{sec:architecture} starts with a short introduction to LSTM RNNs and then describes the single-die \textsc{Muntaniala} architecture and how it can be scaled systolically.
Additionally, that section will describe our demonstrator PCB, called \textsc{Vau da Muntanialas}.
\Cref{sec:results} discusses all results, and \Cref{sec:conclusion} concludes our work.

%%%%%%%%%%%%%%%%%%%%%%%%%%%%%%%%%%%%%%%%%%%%%%%%%%%%%%%%%%%%%%%%%%%%%%%%%%%%%%%%%%%%%%%%
%%%%%%%%%%%%%%%%%%%%%%%%%%%%%%%%%%%%%%%%%%%%%%%%%%%%%%%%%%%%%%%%%%%%%%%%%%%%%%%%%%%%%%%%
\section{Related Work}\label{sec:relatedwork}
%%%%%%%%%%%%%%%%%%%%%%%%%%%%%%%%%%%%%%%%%%%%%%%%%%%%%%%%%%%%%%%%%%%%%%%%%%%%%%%%%%%%%%%%
\subsection{DNN and RNN Acceleration}
In general, the rather complex data dependencies encountered in RNN models make their acceleration more difficult than for feed-forward networks.
Therefore, CPUs and GPUs have difficulties in exploiting the fine grained parallelism encountered in RNNs. 
Even though batching helps with the parallelism, the CPUs and GPUs remain very under-utilized~\cite{Nurvitadhi2016}.

Accelerators implemented on Field-Programmable Gate Array (FPGA) can be more effectively tailored to the requirements of the dataflow of RNNs and, therefore, can achieve a higher energy-efficiency and performance than CPU and GPU implementations.

The FPGA-based accelerator ESE from Han~\etal~\cite{Han2017} works directly on a precompressed parallelization-friendly RNN model.
Their compression method is based on load-balance-aware pruning and compresses the LSTM model by a factor of $10\times$ (without quantization).
Wang~\etal~\cite{Wang2018} apply another compression method based on block-circulant matrices on the model parameters and further reduce the computational complexity with the help of a Fast Fourier Transform algorithm.
Cao~\etal~\cite{Cao2019} introduce Bank-Balanced Sparsity (BBS) and apply it to the model parameters.
Some recent work has been focusing more on the feature maps: The works of~\cite{Cavigelli2018,Gao2018} go a slightly different way and only compute inference on delta-updates, and~\cite{Cavigelli2019} propose a hardware-friendly compression method to reduce the bandwidth requirements to transfer the feature maps. 
However, even with these advanced algorithm adaptations, which sometimes require specialized training methods, the maximum energy-efficiency achieved by FPGA-based accelerators is around 165\,GOP/s/W~\cite{Gao2018}.

However, this energy-efficiency is still too high for energy-constraint embedded IoT devices and highly efficient high-performance computing platforms. FPGAs are, in general, too power-hungry for low-power always-on application scenarios, which require power consumption on the other of a few \si{\milli\watt}.

In the last years, many specialized ASIC accelerators have been proposed for DNN and RNN inference~\cite{Azari2020,Giraldo2018,Kadetotad2020,Yin2018}.
These specialized accelerators achieve two orders of magnitude higher energy-efficiency than the previously presented FPGA-based accelerators.
Many proposed accelerators, such as~\cite{Chen2018,Lee2019,Shin2018}, include, in addition to specialized compute units, one or multiple microcontroller-style cores for dataflow control and computation.
These heterogeneous systems usually focus on accelerating CNNs and fully-connected networks (FCN) instead of RNNs~\cite{Mauro2020}.
In contrast to our proposed \textsc{Muntaniala} design, where the non-linear activation functions are accelerated, these functions are performed on the cores and create a performance bottleneck. They are, therefore, not fully optimized for the complex dataflow dependencies coming with RNNs such as LSTMs.
While accelerators such as EERA-ASR~\cite{Liu2018} and ELSA~\cite{Azari2020} make use of approximate compute units, the accelerator from Kadetotad~\etal~\cite{Kadetotad2020} applies an algorithmic parameter compression technique called hierarchical coarse-grain sparsity (HCGS), which allows reducing the weights by a factor of $16\times$ while keeping the accuracy loss minimal.
\cite{park2018maximizing} proposes a new model parameter compression scheme called compressed and balanced sparse row (CBSR) and shows improved throughput and energy efficiency over the compressed sparse column and rows (CSC and CSR) for an exemplary accelerator placed and routed in 65nm technology.
In contrast, \cite{ardakani2019learning} shows that up to 90\% sparsity can be introduced to the recurrent hidden states without incurring any accuracy degradation on a set of tasks. Their accelerator shows an energy efficiency improvement by up to 5.2$\times$ when zero-skipping these sparse state computations.
All these accelerators are mainly optimized for limited model sizes.
Once the models get bigger than their on-chip storage capacity, they are all forced to reload their parameters, thereby creating an I/O bottleneck which is removed by our Muntaniala design, which, once all parameters are loaded, can work on multiple dies with local on-chip weights.

One way of scaling up the accelerator size is scaling the die size up to a complete wafer, such as the recently presented Cerebras CS-1 wafer-scale engine (WSE)~\cite{Moore2020}.
The Cerebras CS-1 WSE is the largest single die computing system produced so far and comes with benefits such as performance boost by order of magnitude due to the, e.g., lower on-chip communication cost in power and delays compared to on-board communication.
However, the engineering effort and cost needed to produce such a wafer-scale engine are enormous and require not only highly advanced solutions for handling production imperfections but also highly advanced packaging and cooling solutions~\cite{Moore2020}.

Another way to scale up computing systems can be achieved by combining multiple monolithic dies, also called chiplets, within a single package.
Such systems, also called multi-chip-module (MCM), have not only been proposed for high-performance multi-core SoC architectures such as e.g. the works from Vivet~\etal~\cite{Vivet2020} and Zaruba~\etal~\cite{Zaruba2020manticore}, but also for DNN acceleration.
Zimmer~\etal~\cite{Zimmer2019} and Shao~\etal~\cite{Shao2019} proposed to accelerate DNNs with an MCM containing 36 chiplets in a mesh-style network-on-chip (NoC) based communication system using ground-referenced signaling (GRS).
Our \textsc{Muntaniala} prototype includes no advanced I/O interfaces, as no such interface IP  blocks were available for the context of this research.
Nevertheless, our design can be easily adapted to implement any form of MCM, or wafer-scale engine.
Hyperdrive from Andri~\etal~\cite{Andri2019} implements a systolically scalable accelerator very similar to the \textsc{Muntaniala}.
However, they accelerate only CNNs and implement resource-intensive FP16 arithmetic, while we focus on RNNs and work with fixed-point arithmetic.
Additionally, we go one step further and provide not only single prototype results but evaluate the design from a fully fabricated demonstrator including a complete systolic array of $2\times2$ \textsc{Muntaniala} dies.

%%%%%%%%%%%%%%%%%%%%%%%%%%%%%%%%%%%%%%%%%%%%%%%%%%%%%%%%%%%%%%%%%%%%%%%%%%%%%%%%%%%%%%%%
\subsection{Quantization}
Since the rise in demand for DNN inference, many methods have focused on reducing the required numerical precision and in turn mitigate memory bandwidth pressure and compute complexity. With only minimal modifications, DNNs have been shown to run without accuracy loss using reduced bit-width floating-point formats such as IEEE's half-precision (float16), Google's ``brain floating point'' (bfloat16), and Nvidia's 18-bit TensorFloat format---all of which have been implemented on a variety of platforms from microcontrollers to GPUs and application-specific processors~\cite{Kalamkar2019,Ho2017}. 

To further boost the energy-efficiency and performance, several methods have been proposed to enable 8-bit fixed-point inference with merely a calibration phase to optimize the value ranges of the activations and filter weights, with some extracting additional offsets, rescaling factors, or bit-shifts~\cite{Nagel2019}. This has drawn a lot of attention due to the efficiency gains and reduced memory bandwidth for hardware accelerators and its suitability for 4- or 8-way SIMD in processor-based devices. These methods have a very small impact of less than 2\% on the accuracy for most feed-forward networks, although some recent networks such as in MobileNetV2 have been shown to be challenging for re-training-free approaches~\cite{Nagel2019}.

Higher-precision but more computational demanding quantization is based on retraining to adapt network weights to compensate for the quantization effect (quantization-aware training, QAT). This allows to attain almost identical accuracy even for MobileNetV2/V3~\cite{Sandler2018,Howard2019} and almost eliminates the small gap seen for most other networks when quantizing uniformly to 8-bit weights and 8-bit activations. Note that for most of these networks, the (re-)quantization is applied right before the convolutions or after the activations, performing by-pass/skip connections in full precision. Support for QAT is found in many common frameworks such as Keras/TensorFlow or PyTorch and is typically based on the straight-through estimator (STE). 

Besides the common 8-bit QAT, a lot of efforts have been undertaken to further reduce the precision of weights and activations, even down to binary and ternary representations. While in initial efforts, the weights were quantized using STE as well, clear improvements have been shown when a method called \emph{incremental network quantization (INQ)}~\cite{Zhou2017a} was introduced, followed by moderate additional improvements using ADMM and RPR~\cite{Leng2018,cavigelli2020random}. These methods can be applied to uniform quantization but also to power-of-two quantization levels. In contrast, some other methods like LQ-Nets~\cite{Zhang2018a} and TTQ~\cite{Zhu2017} learn the quantization levels to further improve the accuracy at the expense of making their implementation very costly. The commonality of the quantization procedures is their resilience to extreme quantization with 2--3\% accuracy loss for ternary weights and around 1\% for 5-ary weights. 

The feedback loop inherent to RNNs makes them particularly challenging to quantize. In~\cite{He2016b}, they propose a special flavor of STE to quantize weights, balancing their distribution for each layer before quantization during the forward pass and backpropagate the gradients as if there was no quantization. For the quantization of the activations, they perform normal STE, resulting in an overall accuracy drop of 3\% for 4\,bit weights and activations on the IMDB sentence classification dataset. Xu et al.~\cite{Xu2018a} quantize the weights by choosing the nearest quantization level and alternatingly optimize the underlying full-precision weight and the quantization levels, showing an accuracy decrease from 92.5 to 95.2 perplexity per word (PPW) on the PTB dataset. 
Alom et al.~\cite{Alom2018} show an accuracy decrease from 82.9\% to 79.6\% on the IMDB sentiment analysis dataset using an LSTM. Their weight quantization method chooses the quantization levels to have be equi-probably selected, while the activations remain unquantized. 
In~\cite{Hou2019}, Hou et al. elaborate on how additional normalization layers (batch, layer, and weight normalization) generally improve the accuracy of LSTMs on the Penn Treebank dataset for world-level language modeling and show that the weights can be quantized to ternary values without additional loss while keeping the activations in full precision. 
\cite{ardakani2019synthesis} apply stochastic computing to convert all element-wise multiplications into XNOR operations by using binary weights, states, and activations.
They reduce the computational complexity by a factor of 86$\times$ and even improve accuracy over their quantized counterparts across various temporal tasks.

The main focus of this work is the scalable and efficient accelerator architecture.
Therefore, we apply the commonly used QAT methods such as STE for the activations and INQ for the weights in mapping LSTMs to \textsc{Vau da Muntanialas} in \Cref{sec:quant}.

\section{Fundamental Concept: LSTM Networks}
\label{sec:lstm}

Standard Recurrent Neural Networks (RNNs)~\cite{Rumelhart1986} use a feedback of the hidden state $\mathbf{h}_t=(h_1, h_2, ..., h_{N_{H}})$ with $N_{H}$ elements to learn short-time dependencies over time $t=1,...,T$ from an input state $\mathbf{x_{t}}=(x_1, x_2, ..., x_{N_{I}})$ with $N_{I}$ input features iteratively:
\begin{align}
    \mathbf{h}_t = \text{act}(\mathbf{W}_{xh}\mathbf{x}_t+\mathbf{W}_{hh}\mathbf{h}_{t-1}+\mathbf{b}_h) \label{eq:1}
\end{align}
where $\mathbf{W}_{kl}$ are weight matrices, $\mathbf{b}_{l}$ a bias vector whereas the subscripts $k$ stand for the source state/gate ($\mathbf{x}_t$, or $\mathbf{h}_{t-1}$) and the subscripts $l$ for the target state/gate ($\mathbf{h}_t$) to which the source state is contributing to.
The activation function \textit{act()} used in RNNs is typically a non-linear function such as hyperbolic tangent or sigmoidal function.

%%%%%%%%%%%%%%%%%%%%%%%%%%%%%%%%%%%%%%%%%%%%
\begin{figure}[tbp]
\centering
\includegraphics[trim={0 17.2cm 0 0},clip,width=0.5\textwidth]{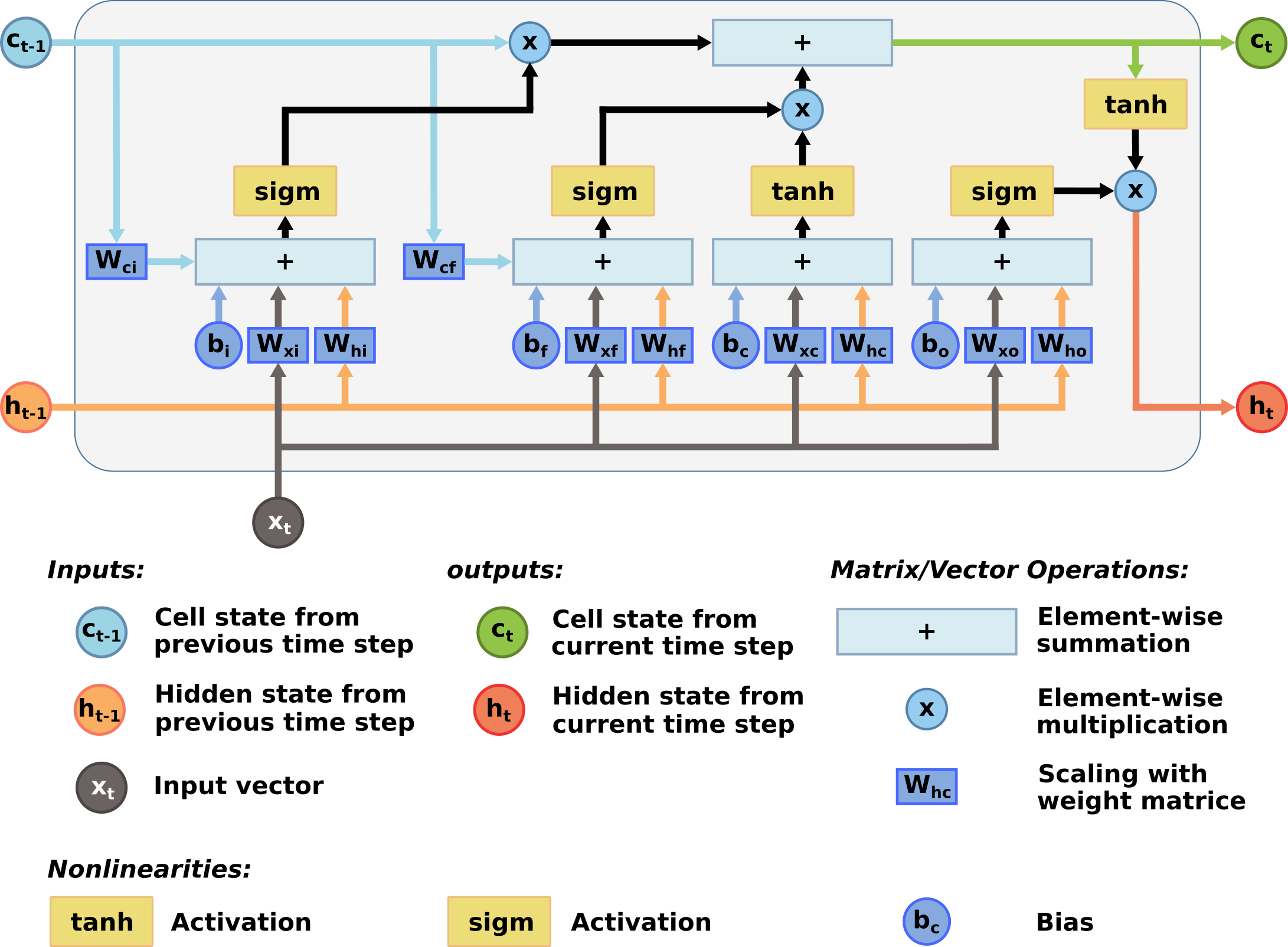}
\caption{Dataflow graph of a peephole LSTM Cell with input vector $\mathbf{x}_{t}$, hidden state $\mathbf{h}_{t}, \mathbf{x}_{t-1}$, and cell state $\mathbf{c}_{t}, \mathbf{c}_{t-1}$.
The compute intensive parts are the vector-matrix multiplications, vector additions, and vector multiplications (blue boxes).
}
\label{fig:lstm}
\end{figure}
%%%%%%%%%%%%%%%%%%%%%%%%%%%%%%%%%%%%%%%%%%%%

Long Short-Term Memory (LSTM) neural networks~\cite{Hochreiter1997}, a special type of RNN, have an additional internal cell state $\mathbf{c}=(c_1, c_2, ..., c_{N_{H}})$ which allows the network to capture not only short-term, but also long-term dependencies. 
Additionally, so-called gates control the information flow to and from the cell state.
An LSTM network layer can be described as follows:
%%%%%%%%%%%%%%%%%%%%%%%%%%%%%%%%%%%%%%%%%%%%
\begin{align}
    \mathbf{i}_{t} &= \sigma( \mathbf{W}_{xi}  \mathbf{x}_t +
                              \mathbf{W}_{hi}  \mathbf{h}_{t-1} +
                              \mathbf{w}_{ci} \odot \mathbf{c}_{t-1} +
                              \mathbf{b}_{i}
                            ), \label{eq:2} \\
    \mathbf{f}_{t} &= \sigma( \mathbf{W}_{xf}  \mathbf{x}_{t} +
                              \mathbf{W}_{hf}  \mathbf{h}_{t-1} +
                              \mathbf{w}_{cf} \odot \mathbf{c}_{t-1} +
                              \mathbf{b}_{f}
                            ), \label{eq:3} \\
    \mathbf{\tilde{c}}_{t} &= \tanh(
                        \mathbf{W}_{xc}  \mathbf{x}_{t} +
                        \mathbf{W}_{hc} \mathbf{h}_{t-1} +
                        \mathbf{b}_{c}
                       ), \label{eq:4}  \\
    \mathbf{c}_{t} &= \mathbf{f}_{t} \odot \mathbf{c}_{t-1} +
                      \mathbf{i}_{t} \odot \mathbf{\tilde{c}}_{t}, \label{eq:5}  \\
    \mathbf{o}_{t} &= \sigma( \mathbf{W}_{xo}\mathbf{x}_{t} +
                              \mathbf{W}_{ho}\mathbf{h}_{t-1} +
                              \mathbf{w}_{co} \odot \mathbf{c}_{t} +
                              \mathbf{b}_{o}
                            ), \label{eq:6} \\
    \mathbf{h}_{t} &= \mathbf{o}_{t} \odot \tanh(\mathbf{c}_{t}), \label{eq:7}
\end{align}
%%%%%%%%%%%%%%%%%%%%%%%%%%%%%%%%%%%%%%%%%%%%
with the input gate $\mathbf{i}$, forget gate $\mathbf{f}$, output gate $\mathbf{o}$, the helper cell candidate state $\mathbf{\tilde{c}}_{t}$ and the cell state $\mathbf{c}$, whereas the subscripts $k$ stand for the source state/gate ($\mathbf{x}_t$, $\mathbf{h}_{t-1}$, or $\mathbf{c}_{t-1}$) and the subscripts $l$ for the target state/gate ($\mathbf{i}_t$, $\mathbf{f}_t$, $\mathbf{c}_t$, $\mathbf{o}_t$) to which the source state is contributing to. 
$\odot$ denotes element-wise multiplication. 
Again, $\mathbf{W}_{kl}$ are weight matrices, $\mathbf{b}_{l}$ are the bias vectors, and the $\mathbf{w}_{kl}$ are peephole weight vectors.
If the weight vectors $\mathbf{w}_{kl} \neq \vec{0}$ the cell state is leaking information from the cell to the gates and the LSTM layer is called peephole LSTM~\cite{Gers2000}. 
However, if the peephole weight vectors $\mathbf{w_ci}$, $\mathbf{w_cf}$, and $\mathbf{w_co}$ are all $\vec{0}$, the layer is called a vanilla LSTM.
\Cref{fig:lstm} shows the dataflow of a peephole LSTM.

The size of such an LSTM layer, also called LSTM cell, is defined by the two parameters $N_{I}$, the number of input features per time step (\# elements in $\mathbf{x}_t$), and $N_{H}$, the number of elements in the hidden and cell state and all the three gates.

LSTM networks can be layered by feeding the hidden state of one LSTM layer/cell as input state) to the the next LSTM cell.
For final processing, the hidden state of the last LSTM layer is often fed into a non-recurrent fully-connected layer (FCL), which can be described as:
%%%%%%%%%%%%%%%%%%%%%%%%%%%%%%%%%%%%%%%%%%%%
\begin{align}
    y_t = \sigma(\mathbf{W}_{y}\mathbf{y}_t+\mathbf{b}_y) \label{eq:8}
\end{align}
%%%%%%%%%%%%%%%%%%%%%%%%%%%%%%%%%%%%%%%%%%%%

%%%%%%%%%%%%%%%%%%%%%%%%%%%%%%%%%%%%%%%%%%%%%%%%%%%%%%%%%%%%%%%%%%%%%%%%%%%%%%%%%%%%%%%%
%%%%%%%%%%%%%%%%%%%%%%%%%%%%%%%%%%%%%%%%%%%%%%%%%%%%%%%%%%%%%%%%%%%%%%%%%%%%%%%%%%%%%%%%
\section{Architecture}\label{sec:architecture}
This section gives an overview of LSTM networks and then introduces the functionality of a stand-alone \textsc{Muntaniala} accelerator.
Afterwards, we describe multi-die systolic scaling, highlight the differences between \textsc{Muntaniala} and its predecessor \textsc{Chipmunk} and finally introduce our demonstrator called \textsc{Vau da Muntanialas}.

%%%%%%%%%%%%%%%%%%%%%%%%%%%%%%%%%%%%%%%%%%%%%%%%%%%%%%%%%%%%%%%%%%%%%%%%%%%%%%%%%%%%%%%%
\subsection{\textsc{Muntaniala}: LSTM Accelerator}
\textsc{Muntaniala} accelerates the inference of peephole LSTM as described by \Cref{lst:pseudo}.
In addition, it can also be configured to run the FCL output layer, often used in LSTM~\cite{AlexGraves2013}.
The core architecture of \textsc{Muntaniala} is derived from \textsc{Chipmunk}~\cite{Conti2018} and mainly differs in the I/O interfaces and, therefore, in the off-die interconnect for the accelerator's systolic setup. 
A summary of these differences can be found later in this section.
\\

%%%%%%%%%%%%%%%%%%%%%%%%%%%%%%%%%%%%%%%%%%%%
\begin{figure}[tbp]
\centering
\includegraphics[clip, trim= 0pt 0pt 0pt 0pt, width=0.46\textwidth]{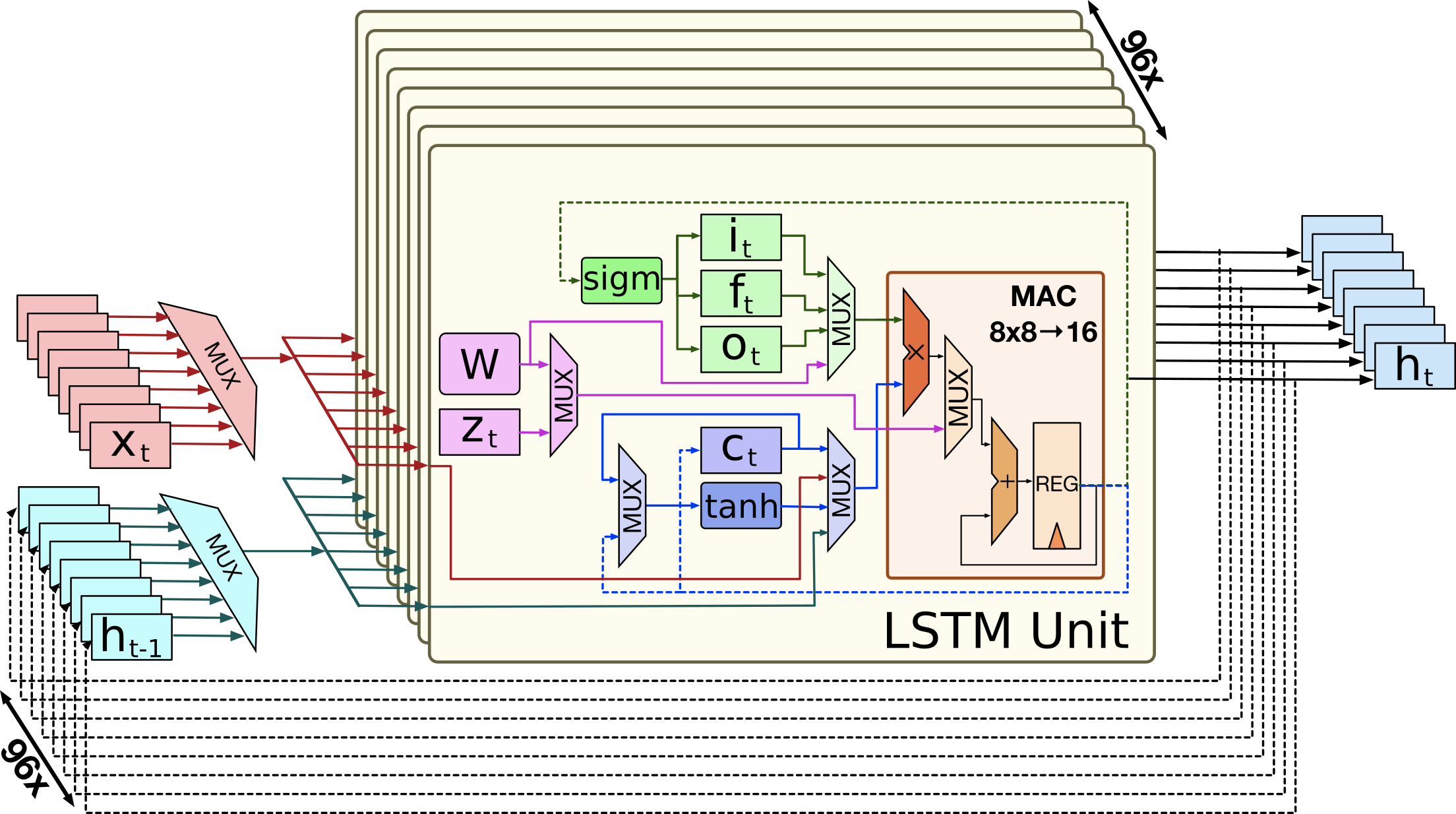}
\centering
\caption{%
LSTM datapath used in \textsc{Chipmunk} and \textsc{Muntaniala}~\cite{Conti2018}.
The datapath implements the operations in \Cref{eq:2,eq:3,eq:4,eq:5,eq:6,eq:7,eq:8}.
}
\label{fig:arch}
\end{figure}
%%%%%%%%%%%%%%%%%%%%%%%%%%%%%%%%%%%%%%%%%%%%

%%%%%%%%%%%%%%%%%%%%%%%%%%%%%%%%%%%%%%%%%%%%%%%%%%%%%%%%%%%%%%%%%%%%%%%%%%%%%%%%%%%%%%%%
\subsubsection{Architecture Overview}\label{subsubsec:arch_overview}

\begin{figure}[tbp]
\centering
\includegraphics[clip, trim= 52pt 422pt 315pt 50pt, width=0.46\textwidth]{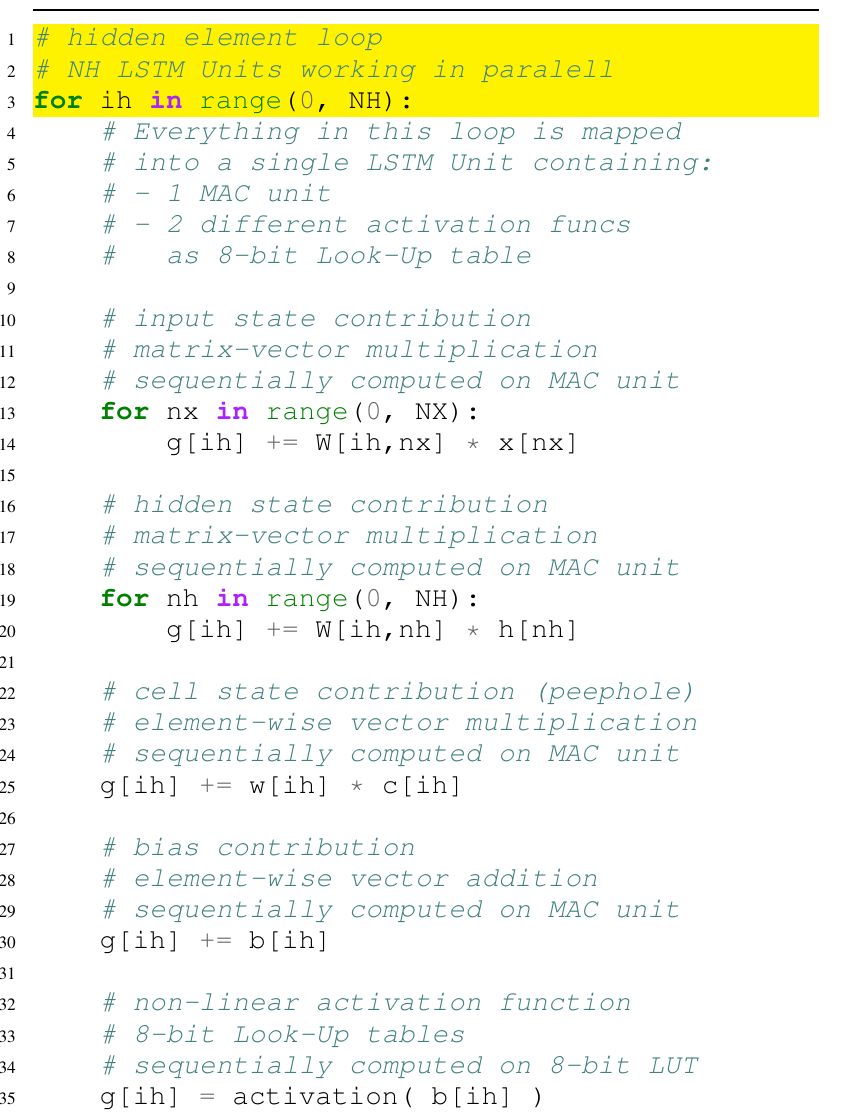}
\centering
\caption{Pseudo-code for a gate or cell-candidate computation within a peephole LSTM layer.}
    \label{lst:pseudo}
\end{figure}

A simplified block diagram of \textsc{Muntaniala}s' datapath and its LSTM units is shown in \Cref{fig:arch}.
The main computational effort of an LSTM inference, as described in \Cref{sec:lstm} comes from the computation of the gates and the cell candidate in \Cref{eq:2,eq:3,eq:4,eq:6} and can be described in a simplified manner by the pseudo-code in \Cref{lst:pseudo}.

%%%%%%%%%%%%%%%%%%%%%%%%%%%%%%%%%%%%%%%%%%%%
%%%%%%%%%%%%%%%%%%%%%%%%%%%%%%%%%%%%%%%%%%%%

An important observation can be made from lines $6-22$ in \Cref{lst:pseudo}:
These computations are mapped as multiply-accumulate operations and are performed sequentially over a single multiply-accumulate unit.
The element-wise non-linear activation function from lines $24-25$ in \Cref{lst:pseudo} are implemented as simple 8-bit look-up tables (LUTs).
The hardware block called LSTM unit includes such a MAC unit and two LUTs, one for $sigmh$ and one for $tanh$.
Additionally, this unit includes some registers for the temporary storage of the gates $\mathbf{i}_t$, $\mathbf{f}_t$, $\mathbf{o}_t$, and cell state $\mathbf{c}_t$, a part of the global SRAM memory for storing all corresponding weights $\mathbf{W}_{i,j}$, $\mathbf{w}_{i}$ and biases $\mathbf{b}_{i}$, and multiplexers for controlling the dataflow.
All states, gates, weights, and biases are stored and processed with $8~\text{bits}$ fixed-point precision while the MAC units make use of $16~\text{bits}$ to minimize overflows.
In \Cref{sec:quant} the choice of quantization and their accuracy implications are further discussed.

\begin{figure*}[tbp]
\centering
\includegraphics[width=0.88\textwidth, trim= 0pt 0pt 0pt 0pt]{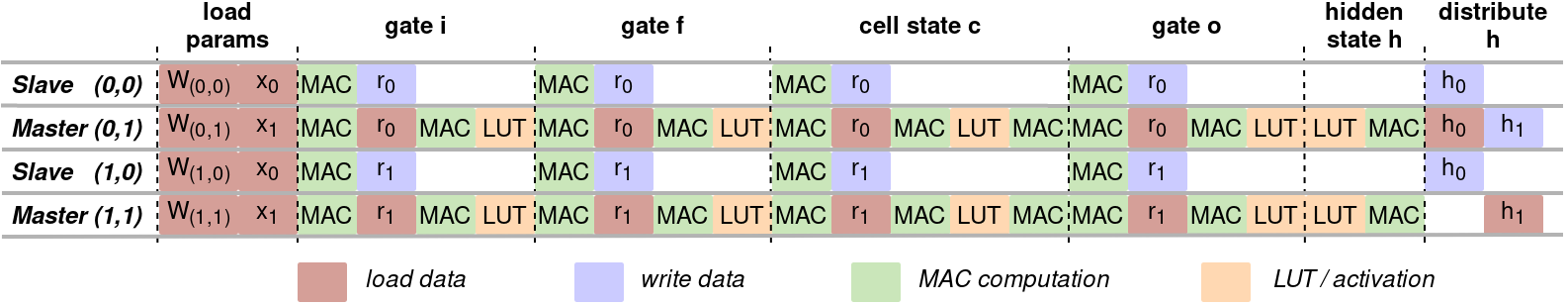}
\caption{The timeline shows the typical computational sequences in a systolic arrangement of $2\times2$ \textsc{Muntaniala} tiles. Note that the block sizes are not proportional to their processing time consumption.
}
\label{fig:timeline}
\end{figure*}

\Cref{fig:timeline} shows the schedule of computation for an example systolic configuration of $2 \times 2$ \textsc{Muntaniala} tiles.
The first (highlighted) hidden element loop on lines $1-3$ in \Cref{lst:pseudo} is computing iteratively every single hidden state element out of all $N_{H}$.
In the \textsc{Muntaniala} architecture, this loop is unrolled, which means the complete accelerator contains $N_{H}$ parallel LSTM Units, where each LSTM Unit works only on one element of the hidden state vector.
For energy-efficiency and optimal performance, \textsc{Muntaniala} loads an initial configuration, all weights, and biases into a local SRAM memory before starting with the computation.
For our prototype with $N_{H}=96$, the total memory is split into 12 SRAM banks to provide enough bandwidth to all $N_{H}$ parallel LSTM Units.
\\

%%%%%%%%%%%%%%%%%%%%%%%%%%%%%%%%%%%%%%%%%%%%
%%%%%%%%%%%%%%%%%%%%%%%%%%%%%%%%%%%%%%%%%%%%

%%%%%%%%%%%%%%%%%%%%%%%%%%%%%%%%%%%%%%%%%%%%
\subsubsection{Systolic Design}\label{sec:systolic}
The architectural design of \textsc{Muntaniala} can be trivially scaled up to support a larger hidden cell size $N_{H}$ per die, which implies increasing the on-chip memory and the number of computational LSTM Units.
However, it is not cost-effective and ultimately infeasible (due to decreasing yield) to arbitrarily increase the die size. 
The \textsc{Muntaniala} design addresses this issue and provides a low-cost solution that scales in a systolic fashion by combining multiple fixed-size dies or chips on an interposer or a circuit board, respectively.

%%%%%%%%%%%%%%%%%%%%%%%%%%%%%%%%%%%%%%%%%%%%
\begin{figure}[tbp]
\centering
\subfloat[][Input State \newline Loading]{\includegraphics[clip, trim= 10pt 50pt 1150pt 20pt, width=0.13\textwidth]{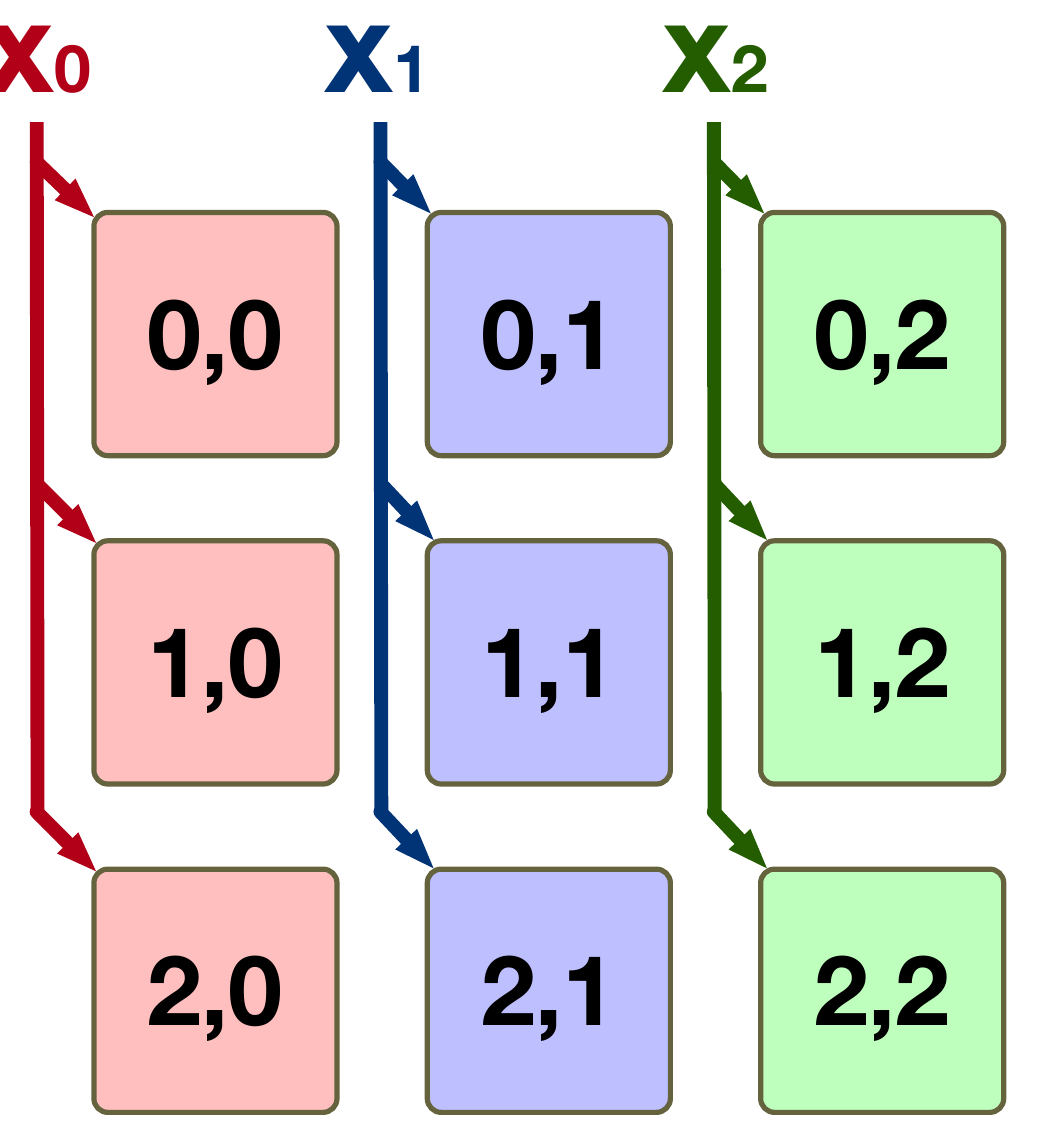}%
\label{fig:input_state}}
\hfil
\subfloat[][Next State \newline Computation / \newline Reduction]{\includegraphics[clip, trim= 350pt 50pt 790pt 20pt, width=0.14\textwidth]{img/systolic.pdf}%
\label{fig:reduction}}
\hfil
\subfloat[][Hidden State \newline Distribution]{\includegraphics[clip, trim= 720pt 50pt 350pt 20pt, width=0.17\textwidth]{img/systolic.pdf}%
\label{fig:hidden_state}}
\newline 
\subfloat[][Inter-Layer Connection]{\includegraphics[clip, angle=0, trim= 0pt 0pt 0pt 0pt, width=0.3\textwidth]{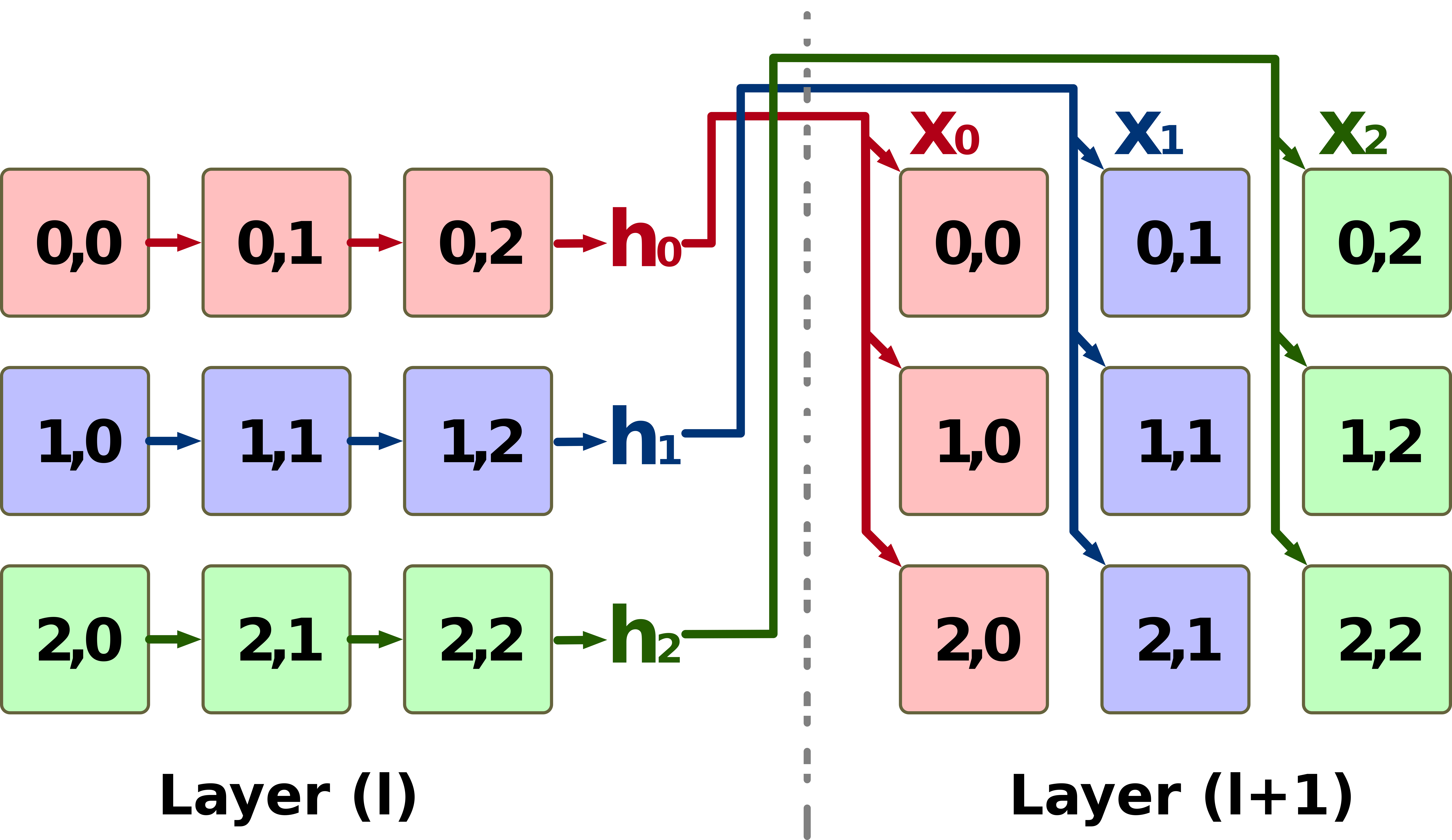}%
\label{fig:multilayer}}
\centering
\caption{%
For accelerating an LSTM network bigger than the network that fits on a single \textsc{Muntaniala} die, multiple dies can be combined together~\cite{Conti2018}.
\Cref{fig:input_state,fig:reduction,fig:hidden_state} show the necessary communication when scaling the size of a single LSTM layer.
\Cref{fig:multilayer} shows how the dies need to be connected for creating multiple LSTM layers.
}
\label{fig:systolic}
\end{figure}
%%%%%%%%%%%%%%%%%%%%%%%%%%%%%%%%%%%%%%%%%%%%

The main computational load of LSTM networks comes from the matrix-vector multiplications (see lines $6-14$ in \Cref{lst:pseudo}), whose problem size scales quadratically with the hidden state's size.
Once the problem size gets too big to fit on a single die, the matrix-vector multiplication can be divided into many smaller multiplications by tiling the weights and vectors and distribute them accordingly on multiple dies, as shown in \Cref{fig:systolic}.
Following the quadratic problem size scaling of the matrix-vector multiplication, scaling the layer size up by $N_{H}= n \times N_{H_{\textsc{Muntaniala}}}$ results in a systolic array of $n \times n = n^{2}$ \textsc{Muntaniala} dies.
By giving every tile at position ($i,j$) the corresponding $\frac{1}{n}$-th tile of the matrix and the corresponding input state tile as in \Cref{fig:input_state}, every die can compute a partial result which needs to be communicated to the next-right die which combines the received and its own partial result, see \Cref{fig:reduction}.
This reduction is performed until the rightmost dies received the results from all other dies in its row, which performs the missing final element-wise activation function.
For the next inference step, every hidden state tile needs to be distributed according to \Cref{fig:hidden_state}.
\Cref{fig:timeline} shows the typical sequences of computations in a systolic arrangement of $2\times2$ \textsc{Muntaniala} dies.
As shown, the non-rightmost dies (also called \emph{slaves}) are stalled.
Simultaneously, the rightmost dies (also called \emph{masters}) are finishing their final computations, typically consisting of accumulations and activations.

Previously, we only considered scaling of a single LSTM layer.
Another possibility are deeper networks by feeding the hidden states output of one LSTM layer as the input state to another LSTM layer.
The \textsc{Muntaniala} design supports this layer stacking by connecting the corresponding data stream interfaces from multiple quadratic grids, as shown in \Cref{fig:multilayer}.
An LSTM network of $l$ layers where each layer has a size of $n \times N_{H_{\textsc{Muntaniala}}}$ would require to combine $l$ grids of $n \times n$ dies, which would in total amount to $l \times n \times n$ dies.

Another possibility of computing multiple layers on fewer dies can be done by not loading the weights only once but reloading them for every layer or sub-tile of a layer.
This means that for $l$ layers, each of size $n \times N_{H_{\textsc{Muntaniala}}}$, only $1\times n \times n$ dies can be used.
After the computation of the first layer, all internal gates $\mathbf{i}_t$, $\mathbf{f}_t$, $\mathbf{o}_t$, and states $\mathbf{c}_t$, $\mathbf{h}_t$ need to be read out and stored externally.
Before the computation of the new layer can start, new parameters (weights and biases) and the corresponding stored gates and states are loaded.
Of course, this has a significant impact on the performance and latency of the complete inference and is, therefore, more reasonable for less latency-critical classifications.
\\

%%%%%%%%%%%%%%%%%%%%%%%%%%%%%%%%%%%%%%%%%%%%
\begin{table}[tbp]
\centering
\caption{I/O Pin Comparison Between \textsc{Chipmunk} and \textsc{Muntaniala}}
\label{tbl:io}
\begin{threeparttable}
\begin{tabular}{@{}lccclccc@{}}
\toprule
\multirow{1}{*}{}   & \multicolumn{3}{c}{\textsc{Chipmunk}}  &  & \multicolumn{3}{c}{\textsc{Muntaniala}} \\ 
\cmidrule(l){2-4} \cmidrule(l){5-8} 
                    & Direction  & \# Data       &  \# HS &  & Direction  &   \# Data    & \# HS   \\ \midrule
Clk / Rst           & I          & 2             & -      &  & I          & 2            & -    \\
Config / Sync       & I          & 1             & -      &  & I          & 3            & -    \\ 
Testing             & I          & 5 (+4)        & -      &  & I          & 2            & -    \\ \midrule
Parameter $p$       & \multicolumn{1}{c}{\multirow{3}{*}{$\Bigg\}$I}} & \multirow{3}{*}{8} & \multirow{3}{*}{2} & & I & 4   & 2    \\
Reduction $r$       &            &               &        &  & I          & 4            & 2    \\
Hidden $h$          &            &               &        &  & I          & 4            & 2    \\ \midrule
Output              & O          & 8             & 2      &  & O          & 4            & 3\tnote{a}    \\ \midrule
\textbf{Total}      & & \multicolumn{2}{c}{\textbf{32}}   &  &            & \multicolumn{2}{c}{\textbf{32}}  \\
\bottomrule
\end{tabular}
\begin{tablenotes}[para, flushleft]
    \item [a] the output interface of \textsc{Muntaniala} has two ready signals allowing to either distribute the hidden state to other dies or the external system, e.g., an FPGA.
\end{tablenotes}
\end{threeparttable}
\end{table}
%%%%%%%%%%%%%%%%%%%%%%%%%%%%%%%%%%%%%%%%%%%%

%%%%%%%%%%%%%%%%%%%%%%%%%%%%%%%%%%%%%%%%%%%%
\begin{figure*}[tbp]
\centering
\subfloat[\textsc{Vau da Muntanialas} Interconnect without chip select]{\includegraphics[clip, angle=0, trim= 0pt 0pt 0pt 0pt,  width=0.4\textwidth]{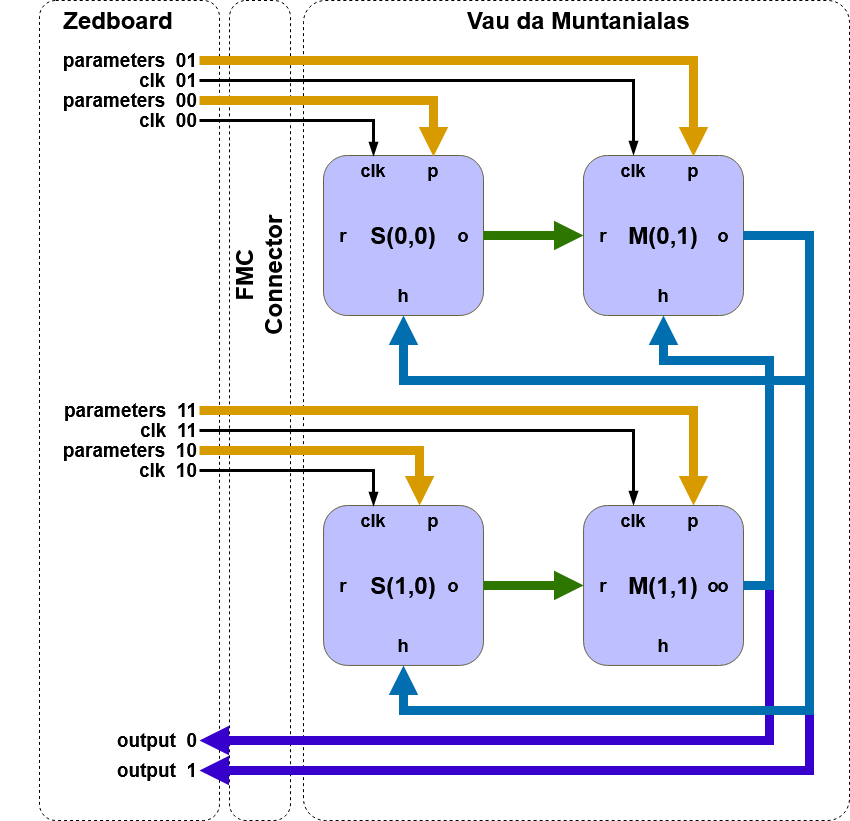}%
\label{fig:pcb_connection}}
\hspace{0.005\textwidth}
\centering
\subfloat[Alternative die connections with chip select]{\includegraphics[clip, angle=0, trim= 0pt 0pt 0pt 0pt,  width=0.4\textwidth]{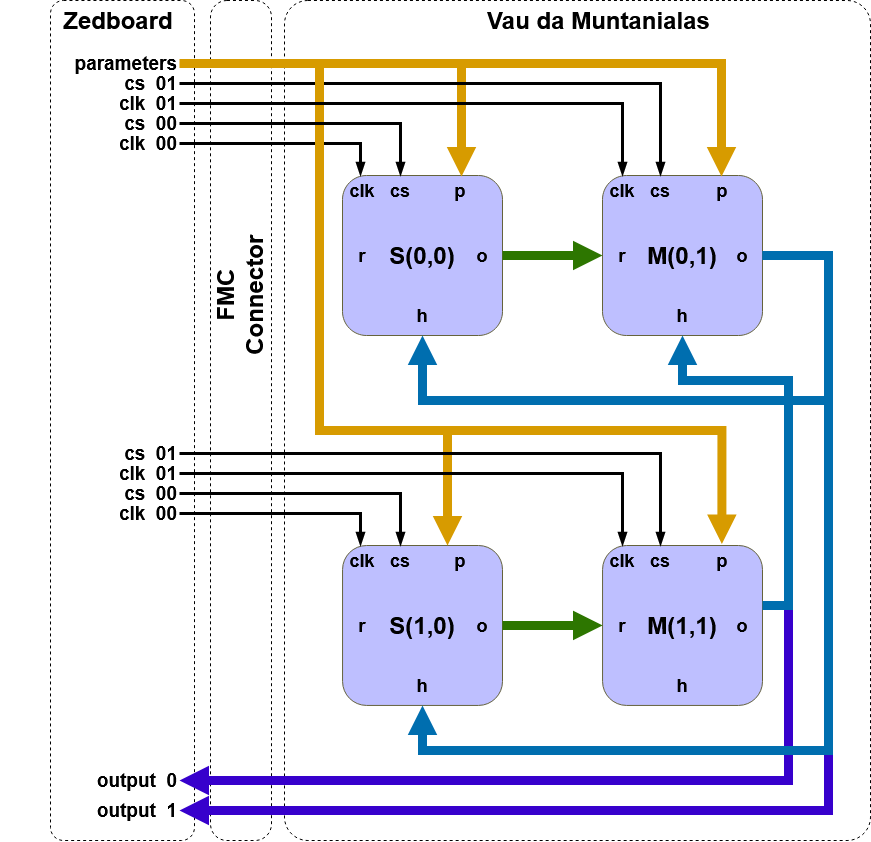}%
\label{fig:cs}}
\centering
\hspace{0.005\textwidth}
\subfloat[Connection legend]{\includegraphics[clip, angle=0, trim= 0pt 0pt 0pt 0pt,  width=0.15\textwidth]{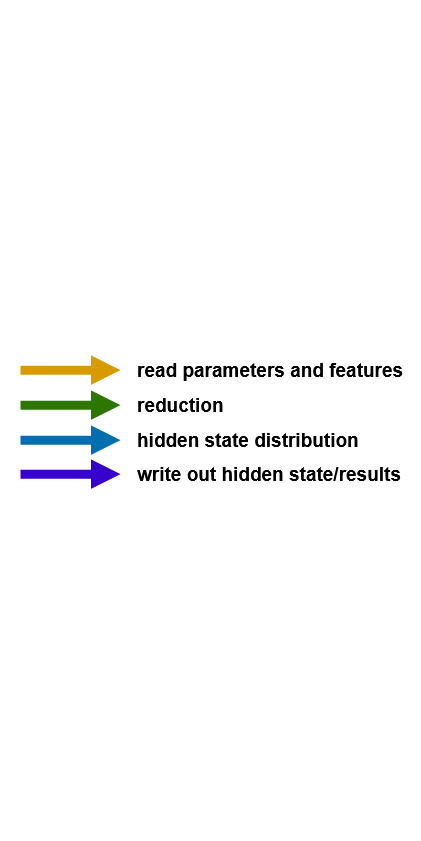}%
\label{fig:legend}}
\caption{%
\Cref{fig:pcb_connection} shows the connection between 2$\times$2 \textsc{Muntaniala} dies on the demonstrator \textsc{Vau da Muntanialas}.
\Cref{fig:cs} shows the connection for an alternative system making us of chipselecet signals for the parameter distribution.
}
\label{fig:2x2}
\end{figure*}
%%%%%%%%%%%%%%%%%%%%%%%%%%%%%%%%%%%%%%%%%%%%

%%%%%%%%%%%%%%%%%%%%%%%%%%%%%%%%%%%%%%%%%%%%width=0.1725\textwidth 
\begin{figure}[tbp]
\centering
\subfloat[\textsc{Chipmunk} Interfaces]{\includegraphics[clip, angle=0, trim= 0pt 0pt 0pt 0pt,  width=0.3\textwidth]{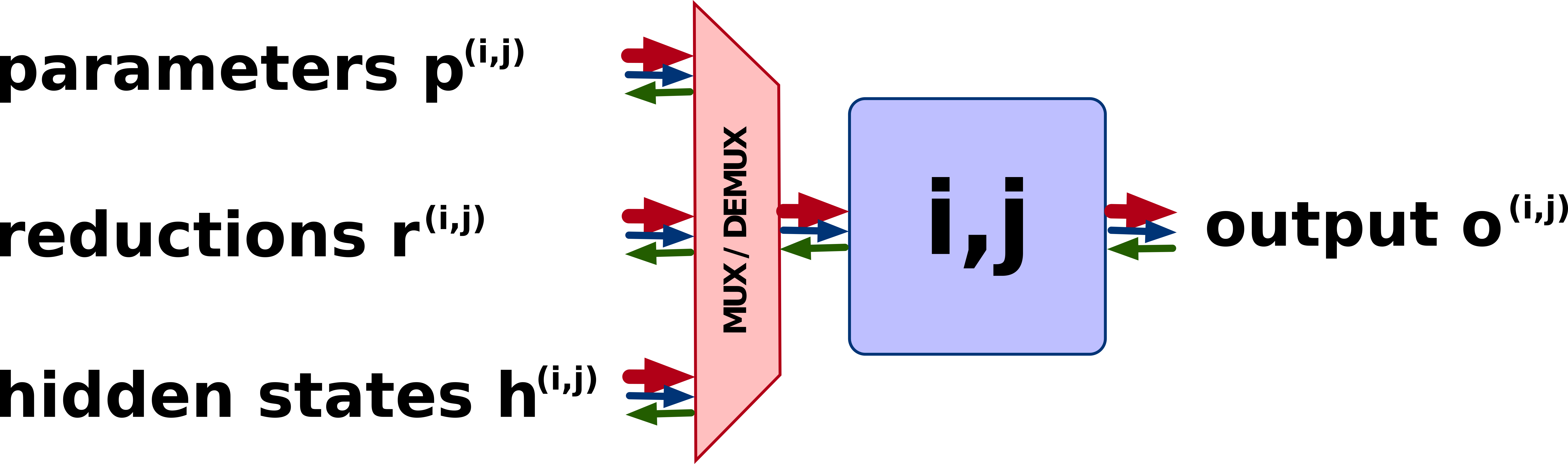}%
\label{fig:c_interconnect}}
\hspace{0.005\textwidth}
\centering
\subfloat[\textsc{Chipmunk} Interface signals]{\includegraphics[clip, angle=0, trim= 0pt 0pt 0pt 0pt,  width=0.13\textwidth]{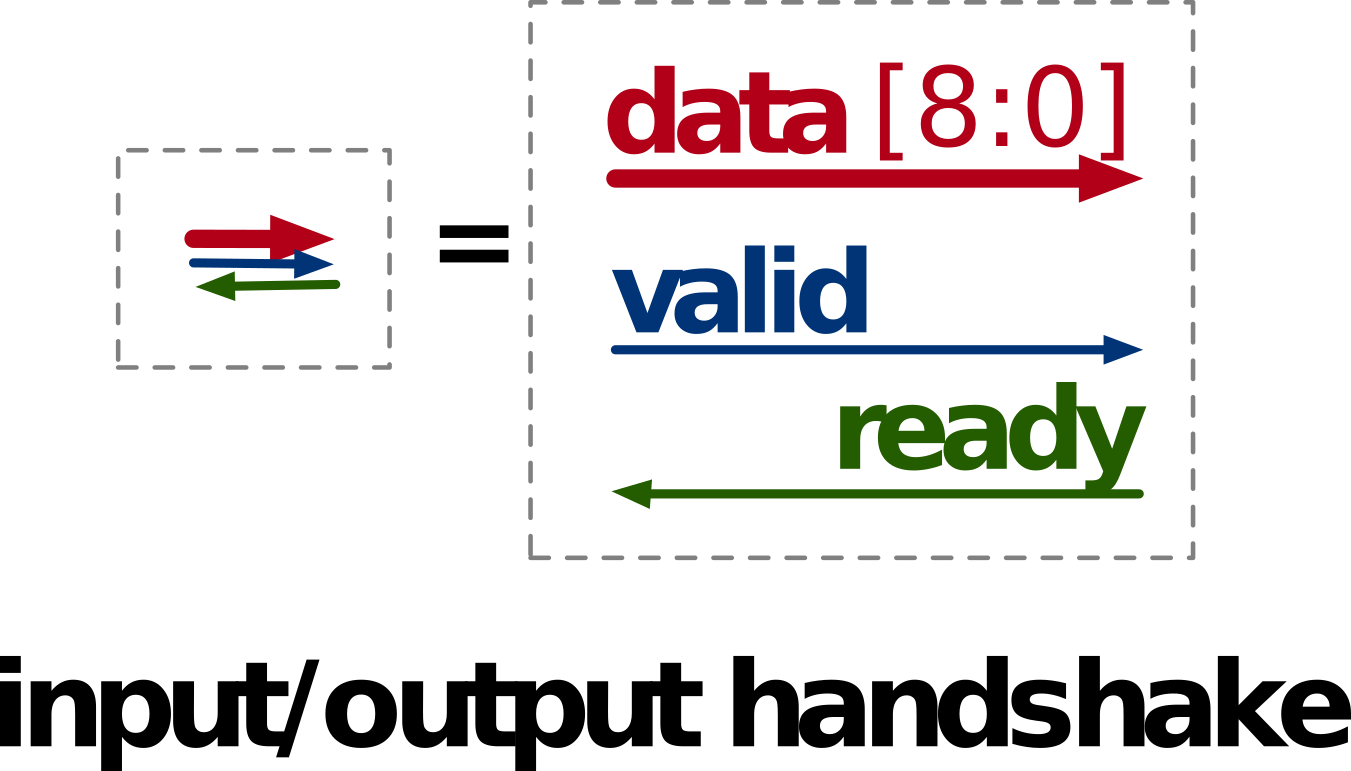}%
\label{fig:c_interconnect_hs}}
\newline
\centering
\subfloat[\textsc{Muntaniala} Interfaces]{\includegraphics[clip, angle=0, trim= 0pt 0pt 0pt 0pt,  width=0.25\textwidth]{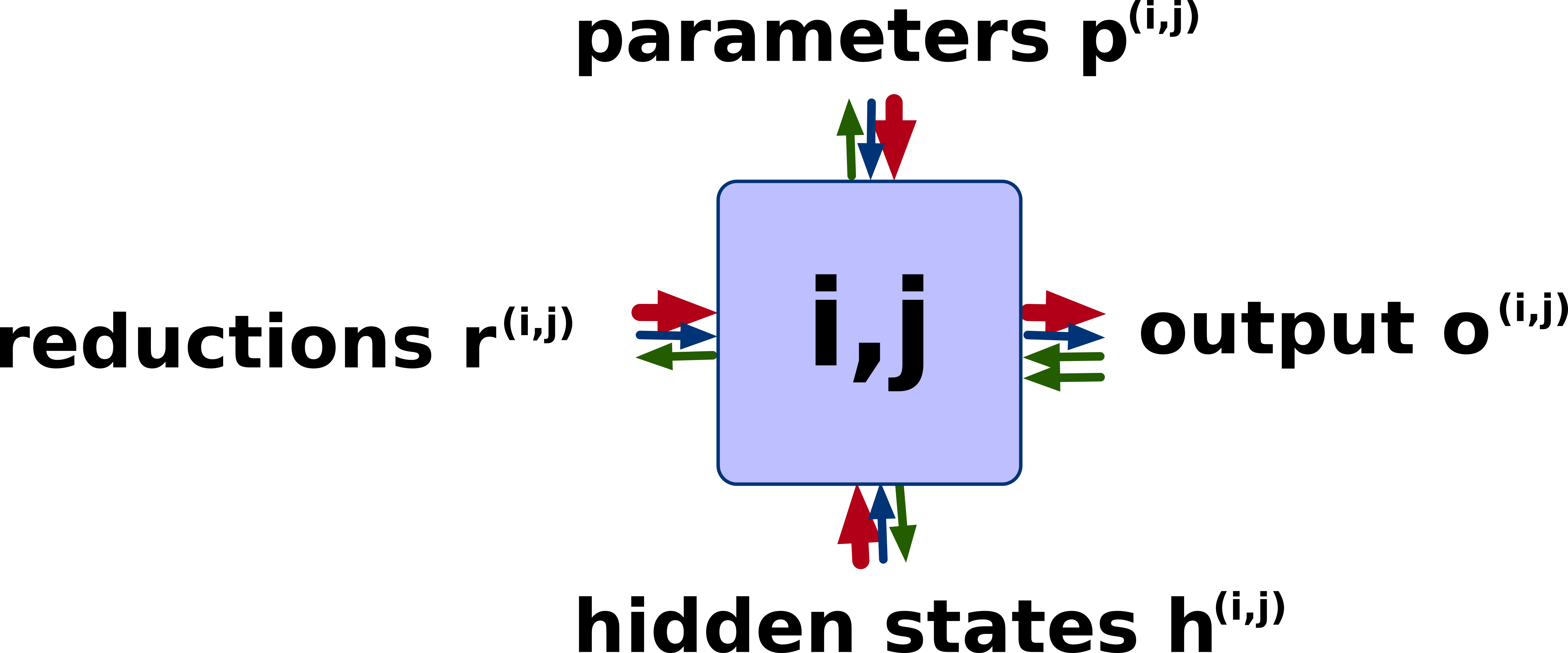}%
\label{fig:m_interconnect}}%
\hspace{0.055\textwidth}
\centering
\subfloat[\textsc{Muntaniala} Interface signals]{\includegraphics[clip, angle=0, trim= 0pt 0pt 0pt 0pt,  width=0.13\textwidth]{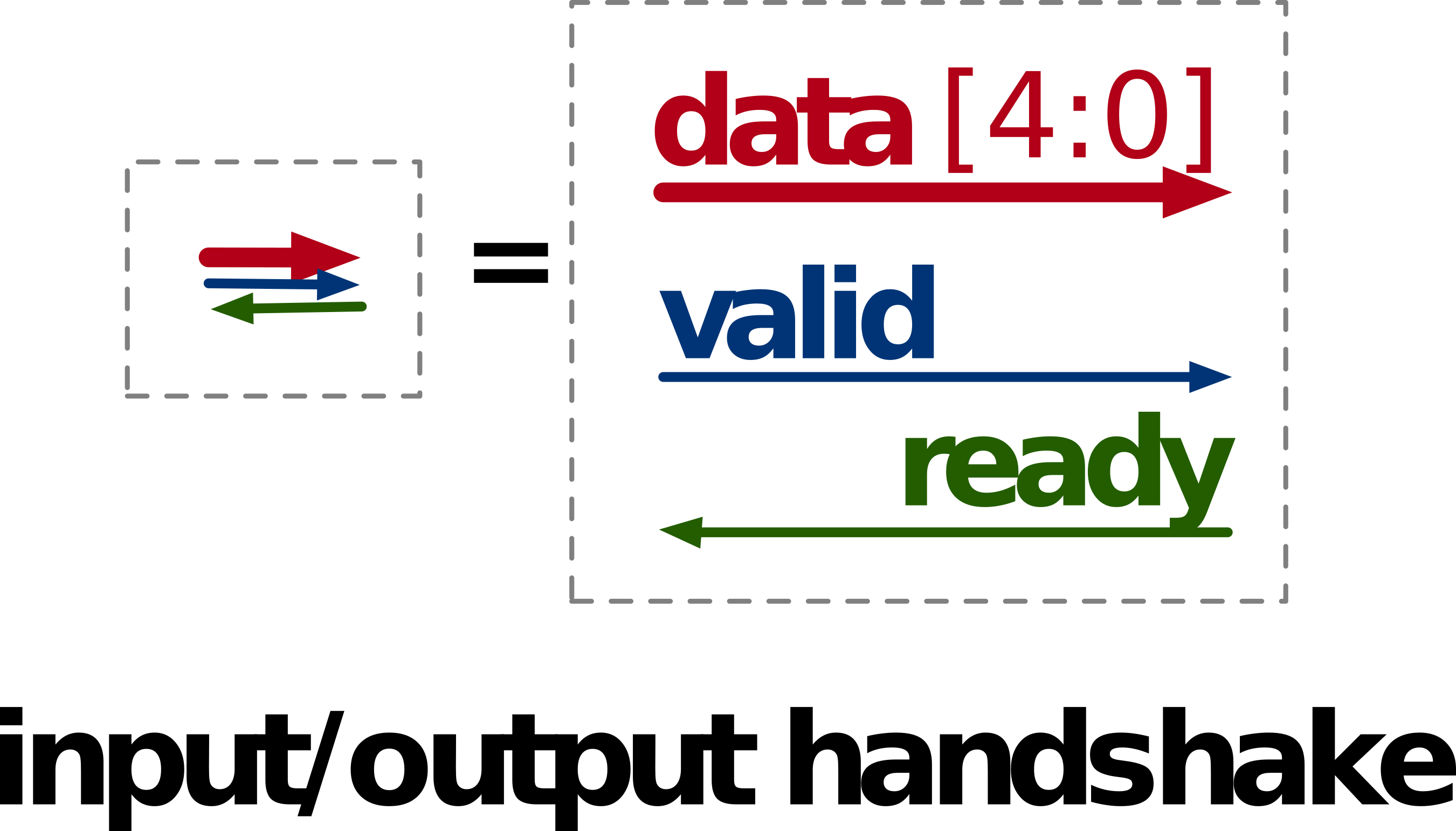}%
\label{fig:m_interconnect_hs}}

\centering
\caption{The different interfaces of \textsc{Chipmunk} and \textsc{Muntaniala}~\cite{Conti2018}.
Note that the output interface of \textsc{Muntaniala} has two ready signals allowing to either distribute the hidden state to other dies or the external system, e.g., an FPGA.
}
\label{fig:interconnect}
\end{figure}
%%%%%%%%%%%%%%%%%%%%%%%%%%%%%%%%%%%%%%%%%%%%

%%%%%%%%%%%%%%%%%%%%%%%%%%%%%%%%%%%%%%%%%%%%
\subsubsection{Improvements: \textsc{Muntaniala} vs \textsc{Chipmunk}} \label{subsubsec:diff}
The main improvement from \textsc{Muntaniala} on its predecessor, \textsc{Chipmunk}, is the new, more compact, and specialized usage of I/O pins.
As shown in \Cref{tbl:io}, every data stream is accompanied by two signals, valid and ready, which together allow communicating via a basic and straightforward handshake protocol.
The only exception is the output interface of \textsc{Muntaniala}, which has two ready signals: one for the handshake with other \textsc{Muntaniala} dies and one for the handshake with the external control unit, e.g., an FPGA.
While \textsc{Chipmunk} has only two bundled data interfaces: one 8\,bit input, and one 8\,bit output data stream, \textsc{Muntaniala} instantiates an interface per transmission source:
\begin{itemize}
    \item \textbf{network parameters $p^{(i,j)}$} such as weights, biases, and new features are fed in from an external controller (e.g., FPGA, microcontroller, ...)
    \item \textbf{intermediate reduction results $r^{(i,j)}$} are received from a neighboring accelerator
    \item \textbf{hidden states $h^{(i,j)}$} for the next inference step are received from a different neighboring accelerator than the reduction results
\end{itemize}
The indices $i,j$ refer to the position of the receiving accelerator.
The usage of three independent data sources allows multiple \textsc{Muntaniala} dies to be connected either on-silicon or on board-level directly, thus reducing design complexity, off-chip load, etc.
The resulting simplified data stream handling is shown in \Cref{fig:m_interconnect}.
However, as during any communication phase, only one input interface is active (or one output interface), the maximum number of data transferred is reduced from 8 bit/cycle to 4 bit/cycle. Correspondingly, the bandwidth is reduced from 168MB/s to 79.5MB/s. Note that the \textsc{Muntaniala} prototype was designed as a proof of concept. For industrial implementation, not only the die size (and with it the hidden state size on a single \textsc{Muntaniala} die) could be scaled up further, but also the used interface could be changed. For example, the High-Bandwidth Interconnect (HBI) offers a high-bandwidth, low-power and low-latency interconnect and is a general standard for die-to-die communication and corresponding IPs are offered by, e.g., Synopsys~\cite{odsa,hbi}.
\\

%%%%%%%%%%%%%%%%%%%%%%%%%%%%%%%%%%%%%%%%%%%%
\subsubsection{System I/O Scaling} \label{subsubsec:ioscale}
Every stand-alone accelerator needs external memory to supply the parameters and input data.
Therefore, the bandwidth and I/O requirements to such an external memory are an essential design aspect for targeting the integration of \textsc{Muntaniala} into an end-to-end system.
From \Cref{tbl:io} it is visible that a single die requires 4 data plus two handshake pins for receiving all parameters and the input features from the environment.
Additionally, every chip has 3 I/O pins, which help with the configuration and synchronization of the dies.
They are controlled externally, signaling the accelerator, e.g., to store out the internal states, load new states, load new parameters, or similar, and can typically be shared over all dies within a systolic array.

An important point to notice is that the I/O for the parameter distribution can be time-multiplexed and be targeted by a simple chip select signal as shown in \Cref{fig:cs}.
Of course, this is better for systolic configurations that do not need to reload their weights often.
Therefore, the minimal required I/O is limited by the number of dies in the input layer $n_{inp\_layer}$ and the number of dies in the output layer $n_{out\_layer}$.
These are the minimal pins needed to feed the input features and write back the final results: $\#IO_{clk,rst}+\#IO_{config} + n_{inp\_layer} \times (4+2) + n_{out\_layer} \times (4+2)$ pins.
At the cost of some additional latency, time-multiplexing for the incoming and outgoing data could be introduced, resulting in a reduced minimum required I/O of $\#IO_{clk,rst} + \#IO_{config} + 1 \times (4+2) + 1 \times (4+2) = 17$ pins.

%%%%%%%%%%%%%%%%%%%%%%%%%%%%%%%%%%%%%%%%%%%%
\begin{figure}[tbp]
\centering
\includegraphics[width=0.5\textwidth]{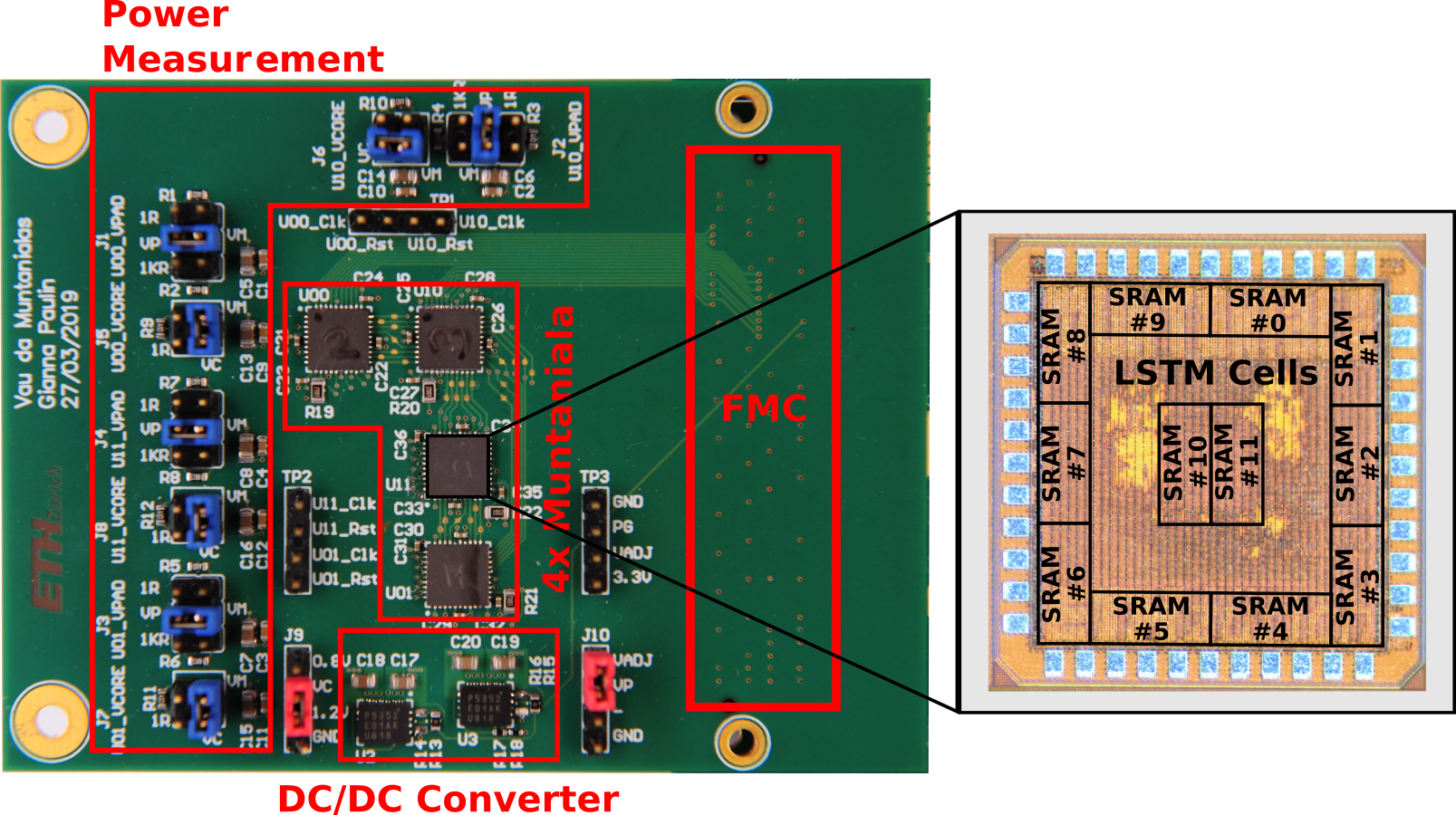}
\caption{The \textsc{Vau da Muntanialas} FMC connector card with a microphotograph of a \textsc{Muntaniala} die.}
\label{fig:vaupcb}
\end{figure}
%%%%%%%%%%%%%%%%%%%%%%%%%%%%%%%%%%%%%%%%%%%%

%%%%%%%%%%%%%%%%%%%%%%%%%%%%%%%%%%%%%%%%%%%%%%%%%%%%%%%%%%%%%%%%%%%%%%%%%%%%%%%%%%%%%%%%
\subsection{Vau da Muntaniala: Systolic Demonstrator}\label{sec:vau}
The aforementioned improvements of the \textsc{Muntaniala} interface allow a direct connection on a Printed Circuit Board (PCB) between the accelerator interfaces in a systolic grid without requiring additional off-chip components such as multiplexers.
We built a demonstrator PCB, called \textsc{Vau da Muntanialas} to take more detailed measurements on the interaction of multiple \textsc{Muntaniala} dies.

This PCB is controlled using a simple Field Programmable Gate Array (FPGA) \footnote{\url{http://zedboard.org}}.
The Zynq-7000 system-on-chip (SoC) combines tightly coupled ARM cores with an FPGA.
The FMC connector allows us to control our custom FMC card, i.e., our demonstrator PCB, with the Zynq FPGA.
\Cref{fig:vaupcb} shows our custom FMC card including the grid of $2\times 2$ \textsc{Muntaniala} LSTM accelerator chips, capable of collaboratively performing LSTM inference with $1$ layer with a hidden state size of $2\times N_{H_{\textsc{Muntaniala}}} = 192$.
The four \textsc{Muntaniala} dies are interconnected as shown in \Cref{fig:pcb_connection}.
\\

%%%%%%%%%%%%%%%%%%%%%%%%%%%%%%%%%%%%%%%%%%%%
\subsubsection{FMC Card}
As shown in \Cref{fig:vaupcb}, the FMC card includes not only the four \textsc{Muntaniala} accelerators but also two DC/DC converters, multiple power measurement points, and an FPGA Mezzanine Card (FMC) connector.
For more fine-grained control and evaluation, the PCB can also be powered from an external power supply.
The four \textsc{Muntaniala} accelerators were arranged as an upside-down letter ``L'' to minimize the potential interference caused by unbalanced board connections.
The clocks and resets of all accelerators could, if needed, be controlled individually. However, this was not necessary during our evaluation.
\\

%%%%%%%%%%%%%%%%%%%%%%%%%%%%%%%%%%%%%%%%%%%%
\subsubsection{FPGA-Board Implementation}
The processing system (PS) part of the Zynq reads parameters from the SD-Card.
Over an AXI interconnect, the simple bare-metal implementation writes the parameters into the BRAM on the programmable logic (PL) part of the Zynq.
Once all parameters are stored in the BRAM, the PS notifies a custom HDL controller over the same AXI interconnect.
The controller is implementing the same basic handshaking protocol of \textsc{Muntaniala} as explained in \Cref{subsubsec:diff} and transfers the BRAM data over the FPGA's programmable I/Os, which are connected to the corresponding FMC pins.

%%%%%%%%%%%%%%%%%%%%%%%%%%%%%%%%%%%%%%%%%%%%
\begin{table}[t]
\caption{Training Hyperparameters for the LSTM Network on the TIMIT Dataset}
\label{tbl:hyperparam}
\centering
\begin{tabular}{ll}
\toprule
Optimizer              & Adam   \\
Learning Rate          & 0.02   \\
\midrule
INQ-Optimizer          & SGD    \\
Learning Rate          & 0.002  \\
\midrule
Loss function          & Pytorch CTC   \\
Batch Norm             & No   \\
Batch Size             & 128  \\
\# Phonemes Training   & 62   \\
\# Phonemes Evaluation & 39   \\
\bottomrule
\end{tabular}
\end{table}
%%%%%%%%%%%%%%%%%%%%%%%%%%%%%%%%%%%%%%%%%%%%

%%%%%%%%%%%%%%%%%%%%%%%%%%%%%%%%%%%%%%%%%%%%
\begin{table}[t]
\caption{Phoneme Error Rates (PER) of various Networks on the TIMIT Dataset}
\label{tbl:per}
\centering
\begin{tabular}{lccc}
\toprule
Network size        & Quantization   & Quantization & PER                      \\
$L$-$N_H$-$N_I$     & Weight         & Activation   & Test Set                 \\
\midrule
3L-384NH-123NI       & FP             & FP           & $26.7$\% \\
3L-384NH-123NI       & 8bit lin. INQ  & 8bit STE     & $30.4$\% \\
\bottomrule
\end{tabular}
\end{table}
%%%%%%%%%%%%%%%%%%%%%%%%%%%%%%%%%%%%%%%%%%%%

%%%%%%%%%%%%%%%%%%%%%%%%%%%%%%%%%%%%%%%%%%%%%%%%%%%%%%%%%%%%%%%%%%%%%%%%%%%%%%%%%%%%%%%%
%%%%%%%%%%%%%%%%%%%%%%%%%%%%%%%%%%%%%%%%%%%%%%%%%%%%%%%%%%%%%%%%%%%%%%%%%%%%%%%%%%%%%%%%
\section{Results}\label{sec:results}

%%%%%%%%%%%%%%%%%%%%%%%%%%%%%%%%%%%%%%%%%%%%%%%%%%%%%%%%%%%%%%%%%%%%%%%%%%%%%%%%%%%%%%%%
\subsection{Evaluation - Quantization of LSTM}\label{sec:quant}

As part of this work, we show that 8\,bit quantization of both the activations and the weights is possible for LSTMs for phoneme recognition. We combine the straight-through estimator (STE) for the activations, i.e. $y=\mathrm{quant}(x)$ in the forward pass and $\delta x = \delta y$ in the backward pass where \emph{quant} maps $x$ to the closest quantization level, with incremental network quantization (INQ)~\cite{Zhou2017a} for the weights. We perform uniform quantization for both, as learned quantization levels would require to decompress the values into a higher precision representation, leading to larger and less energy-efficient compute units. In turn, this excludes more recent quantization methods such as LQ-Nets or TTQ, whose gains are based on learning the quantization levels.

We first fully train the network in high precision before collecting value range statistics and beginning retraining with 255-level STE enabled for all post-activation feature maps and the input features of the network. After convergence, we set the 255 weight quantization levels uniformly across the value range of the weights and start applying INQ. We iteratively increase the share of quantized weights on a 
40\%, 60\%, 80\%, 90\%, 100\%
schedule after convergence for the previous share, quantizing them from largest to smallest magnitude.

The feature extraction computes 40 MFCC features plus 1 energy signal on $25$ \si{\milli\second} of audio every $10$ \si{\milli\second}. Additionally, the derivatives are taken into account resulting in an input feature vector size of $N_X=123$.
The networks were trained with the Connectionist Temporal Classification (CTC) loss for all 62 phonemes.
For the evaluation, the phonemes are merged to the typically used 39 phonemes~\cite{Lee1989} and evaluated with a greedy decoder.
\Cref{tbl:hyperparam} summarizes the hyperparameters and quantization schemes used to train the networks.
\Cref{tbl:per} shows the achieved accuracy on various network sizes.
The quantization only imposes a PER drop of approx. 3\% on a network of the size 3L-384NH-123NI (similar to Graves \etal~\cite{AlexGraves2013}).

For the PyTorch implementation, we use the LSTM class. Therefore, the inference results from the quantized PyTorch network implementation and the Muntaniala accelerator can be slightly different, as our  Look-Up tables implementation takes an 8bit input value. In contrast, the Pytorch implementation takes a full-precision input for the activation function. 
We report a mean squared error (MSE) of $2.965 \times 10^{-5} (\pm 3.691 \times 10^{-5})$ with a maximum squared error of $1.92 \times 10^{-4}$ for the $tanh()$ and a MSE of $2.229 \times 10^{-5} (\pm 2.177 \times 10^{-5})$ with maximum squared error of $8.57 \times 10^{-5}$ for $sigmh()$.
These small errors confirm that the LUT activation error is not a major concern for the inference accuracy on the accelerator.

%%%%%%%%%%%%%%%%%%%%%%%%%%%%%%%%%%%%%%%%%%%%

\setlength{\tabcolsep}{4pt}
\begin{table*} [ht!]
% \small
% \scriptsize{
\footnotesize{
\centering
\caption{Comparison to Existing VLSI Implementations}
\label{tbl:hw_comp}

\begin{threeparttable}
    \begin{tabular}{@{}llccrrrrrrrrrr@{}}
    \toprule
    \multicolumn{1}{l}{\multirow{3}{*}{Publication}} &
    \multicolumn{1}{l}{\multirow{3}{*}{\begin{tabular}[c]{@{}c@{}}Supported\\ Networks\end{tabular}}} &
    \multicolumn{1}{c}{\multirow{3}{*}{Type}} &
    \multicolumn{1}{c}{\multirow{3}{*}{\begin{tabular}[c]{@{}c@{}}Tech. \\ ~[nm]\end{tabular}}} & 
    \multicolumn{1}{c}{\multirow{3}{*}{\begin{tabular}[c]{@{}c@{}}Area \\ ~[mm$^2$]\end{tabular}}} & 
    \multicolumn{1}{c}{\multirow{3}{*}{\begin{tabular}[c]{@{}c@{}}Memory \\ SRAM \\ ~[kB]\end{tabular}}} &
    \multicolumn{1}{c}{\multirow{3}{*}{\begin{tabular}[c]{@{}c@{}}Quant.\\ ~[bit] \end{tabular}}} & 
    \multicolumn{1}{c}{\multirow{3}{*}{\begin{tabular}[c]{@{}c@{}} Nr. \\ of \\ MAC\end{tabular}}} &
    \multicolumn{1}{c}{\multirow{3}{*}{\begin{tabular}[c]{@{}c@{}}Voltage\\ ~[V]\end{tabular}}} & 
    \multicolumn{1}{c}{\multirow{3}{*}{\begin{tabular}[c]{@{}c@{}}Frequ.\\ ~[MHz]\end{tabular}}} & 
    \multicolumn{1}{c}{\multirow{3}{*}{\begin{tabular}[c]{@{}c@{}}Power\\ ~[mW]\end{tabular}}} &
    \multicolumn{1}{c}{\multirow{3}{*}{\begin{tabular}[c]{@{}c@{}}Perf.\\ ~[GOP/s]\end{tabular}}} &
    \multicolumn{1}{c}{\multirow{3}{*}{\begin{tabular}[c]{@{}c@{}}Energy\\ Eff.\\ ~$[\frac{\text{TOP/s}}{\text{W}}]$\end{tabular}}} &
    \multicolumn{1}{c}{\multirow{3}{*}{\begin{tabular}[c]{@{}c@{}}Area\\ Eff.\\ ~$[\frac{\text{GOP/s}}{\text{mm}^2}]$\end{tabular}}} \\
    \multicolumn{1}{c}{} & \multicolumn{1}{c}{} & \multicolumn{1}{c}{} & \multicolumn{1}{c}{} & \multicolumn{2}{c}{} & \multicolumn{1}{c}{} & \multicolumn{1}{c}{} & \multicolumn{1}{c}{} & \multicolumn{1}{c}{} & \multicolumn{1}{c}{} & \multicolumn{1}{c}{} & \multicolumn{1}{c}{} & \multicolumn{1}{c}{}  \\
    \multicolumn{1}{c}{} & \multicolumn{1}{c}{} & \multicolumn{1}{c}{} & \multicolumn{1}{c}{} & \multicolumn{1}{c}{} & \multicolumn{1}{c}{} & \multicolumn{1}{c}{} & \multicolumn{1}{c}{} & \multicolumn{1}{c}{} & \multicolumn{1}{c}{} & \multicolumn{1}{c}{} & \multicolumn{1}{c}{} & \multicolumn{1}{c}{} & \multicolumn{1}{c}{} \\ 
    \midrule
    ELSA~\cite{Azari2020} & LSTM & Si & 65 & 2.62 & 106 & 8-11 & 772 & 1.1 & 322 & 20.4 & 27.0 & 1.32 & 10.3 \\
    Laika~\cite{Giraldo2018} & LSTM FC & Si & 65 & 1.03\tnote{d} & 32 & 8 (32) & 8 & 0.575\tnote{e} & 0.25 & 0.005 & 0.004 & 0.822 & - \\
    Kadetotad \etal~\cite{Kadetotad2020} & LSTM & Si & 65 & 7.74\tnote{d} & 297 & 6,13 & 64 & 1.1 & 80 & 67.3 & \textbf{164.95} & 2.45 & - \\
    Kadetotad \etal~\cite{Kadetotad2020} & LSTM & Si & 65 & 7.74\tnote{d} & 297 & 6,13 & 64 & 0.68 & 8 & 1.85 & 24.6 & \textbf{8.93} & - \\
    OCEAN~\cite{Chen2018} & GRU & Si & 65 & 10.15\tnote{d} & 64 & 16 & 32 & 1.2 & 400 & 155.8 & \textbf{311.6} & 2.0 & - \\
    OCEAN~\cite{Chen2018} & GRU & Si & 65 & 10.15\tnote{d} & 64 & 16 & 32 & 0.8 & 20 & 6.6 & 15.58 & 2.36 & - \\
    Yin \etal~\cite{Yin2018} & CNN FC RNN & Si & 65 & 14.4 & 348 & 8/16 & 512/256 & 1.2 & 200 & 386 & \textbf{409.6} & 1.06 & 28.35 \\
    Yin \etal~\cite{Yin2018} & CNN FC RNN & Si & 65 & 14.4 & 348 & 8/16 & 512/256 & 0.67 & 10 & 4 & 20.4 & \textbf{5.09} & - \\
    DNPU~\cite{Shin2018} & CNN FC RNN & Si & 65 & $\sim$2.00 & 10 & 4-7\tnote{i} & 64 & 1.2 & 200 & 21 & 25 & 1.1 & $\sim$28.39 \\
    DNPU~\cite{Shin2018} & CNN FC RNN & Si & 65 & $\sim$2.00 & 10 & 4-7\tnote{i} & 64 & 0.77 & 50 & 2.6 & 6.25 & 2.40 & - \\
    UNPU~\cite{Lee2019} & CNN FC RNN & Si & 65 & 16.00\tnote{d} & 256 & 1-16\tnote{h} & - & 1.1 & 200 & 297 & \textbf{891.2} & 2.5\tnote{g} & - \\
    UNPU~\cite{Lee2019} & CNN FC RNN & Si & 65 & 16.00\tnote{d} & 256 & 1-16\tnote{h} & - & 0.63 & 5 & 3.2 & 22.28 & \textbf{5.32}\tnote{g} & - \\
    \midrule
    Wu \etal~\cite{Wu2019} & LSTM & P\&R & 40 & 0.45 & 88.5 & 8-16 & 12 & 1.1 & 200 & 6.16 & 24 & 3.89 & \textbf{53.3} \\
    Wu \etal~\cite{Wu2019a} & BLSTM CNN & Synth & 40 & 1.40\tnote{d} & 186 & 8-16 & 16 & 1.1 & 100 & 2.13 & 7.49 & 3.52 & - \\
    \midrule
    AIDA~\cite{Yavits2018} & CNN RNN &  Scaled\tnote{f} & 28\tnote{f} & 44.50 & 6400 & 16 & - & - & 1000 & 7150.0 & \textbf{1474} & 0.206 & - \\
    iFPNA~\cite{Chen2018a} & CNN RNN & Si & 28 & 2.52\tnote{d} & 84 & 4-16 & 348 & 1.16 & 125 & 39.4 & 48\tnote{a} & - & - \\
    iFPNA~\cite{Chen2018a} & CNN RNN & Si & 28 & 2.52\tnote{d} & 84 & 4-16 & 348 & 0.63 & 30 & 3.1 & 11.52\tnote{a} & 0.85 & - \\
    EERA-ASR~\cite{Liu2018} & BWN LSTM & Synth & 28 &  0.32\tnote{d} & $>$56 & 2-16 & 16 & 0.8 & 400 & 54.0 & 179.2 & 3.318 & - \\
    \midrule
    \textsc{Chipmunk}~\cite{Conti2018} & LSTM & Si & 65 & 0.93 & 84 & 8 (16) & 96 & 1.24 & 168 & 29.03 & 32.3 & 1.11 & \textbf{34.4} \\
    \textsc{Chipmunk}~\cite{Conti2018} & LSTM & Si & 65 & 0.93 & 84 & 8 (16) & 96 & 0.75 & 20 & 1.24 & 3.08 & 3.08 & - \\
    \midrule
    This Work & LSTM & Si & 65 & 0.93 & 84 & 8 (16) & 96 & 1.275 & 159 & 30.36 & 30.53 & 1.01 & \textbf{32.8} \\
    This Work & LSTM & Si & 65 & 0.93 & 84 & 8 (16) & 96 & 0.7 & 3.8 & 0.22 & 0.73 & 3.28 & - \\
    \midrule
    This Work & LSTM & Scaled\tnote{c} & 40 & - & 84 & 8 (16) & 96 & 1.1 & 159 & 13.91 & 30.53 & 2.19 & - \\
    This Work & LSTM & Scaled\tnote{c} & 40 & - & 84 & 8 (16) & 96 & 0.7 & 3.8 & 0.135 & 0.73 & 5.4 & - \\
    \bottomrule
    \end{tabular}
    \begin{tablenotes}[para, flushleft]
        \item [a] Peak performance.
        \item [b] To the best of our understanding.
        \item [c] Scaled, using the simple model $\tilde{P}=P(l_{new}/l_{old})(V_{dd,new}/V_{dd,old})^2$.
        \item [d] Die area as the core area is not available.
        \item [e] SRAM at 0.7V.
        \item [f] Design was first synthesized in 45nm and then scaled to 28nm, only 28nm results are available.
        \item [g] To the best of our knowledge this applies only to the FC part of RNN and not to the elementwise multiplication, or non-linear activations which are both performed on a SIMD core.
        \item [h] The numbers correspond for 8bit configuration.
        \item [i] The numbers correspond for 4bit configuration.
    \end{tablenotes}
\end{threeparttable}}
\end{table*}

\setlength{\tabcolsep}{6pt}

%%%%%%%%%%%%%%%%%%%%%%%%%%%%%%%%%%%%%%%%%%%%

%%%%%%%%%%%%%%%%%%%%%%%%%%%%%%%%%%%%%%%%%%%%%%%%%%%%%%%%%%%%%%%%%%%%%%%%%%%%%%%%%%%%%%%
\subsection{\textsc{Muntaniala} - LSTM Accelerator}\label{sec:resacc}
\begin{figure}[tbp]
\centering
\includegraphics[width=0.4\textwidth, trim= 0pt 0pt 0pt 0pt,]{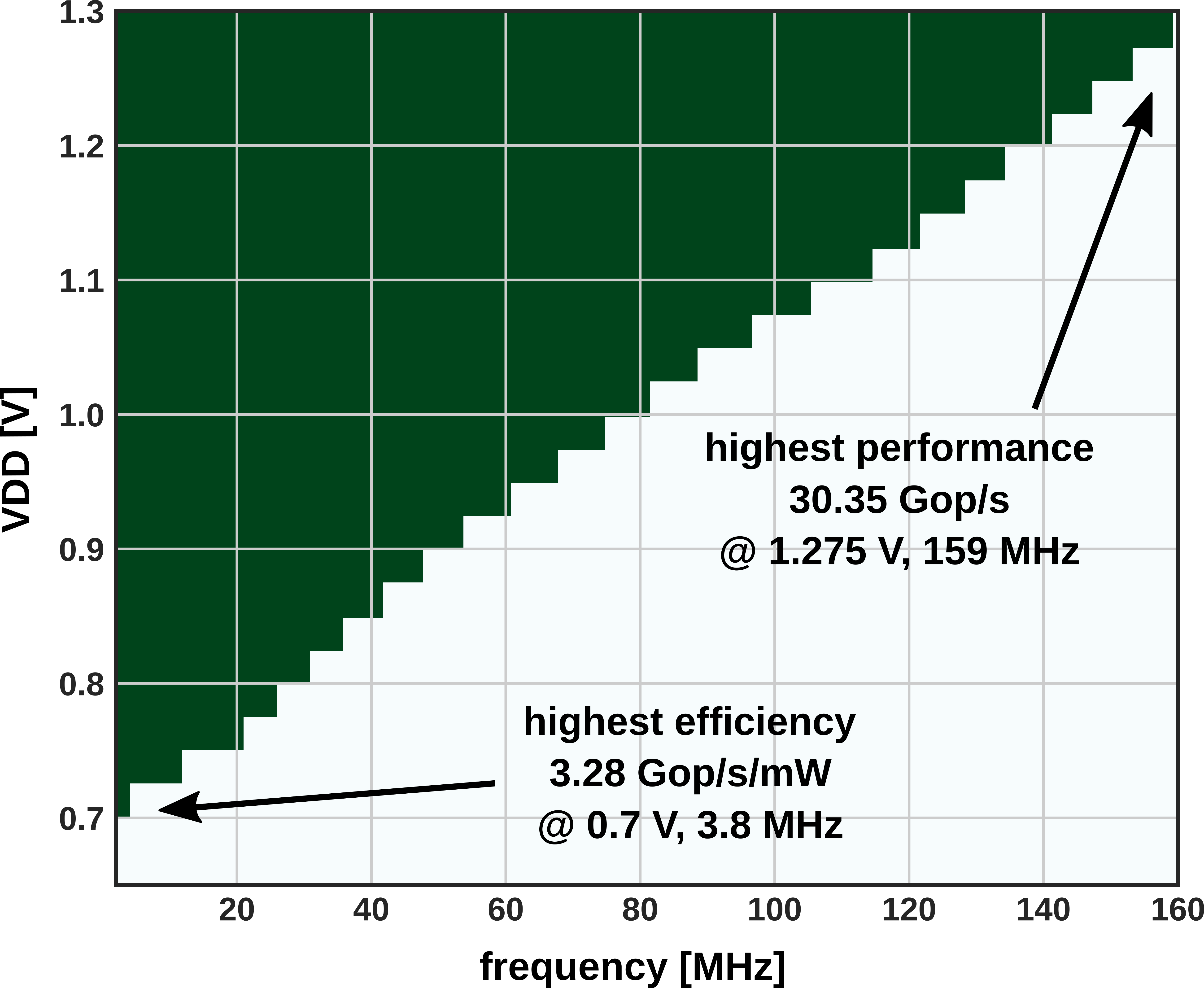}
\caption{\textsc{Muntaniala} Shmoo Plot.}
\label{fig:shmoo}
\end{figure}

The silicon prototype of a single \textsc{Muntaniala} tile as described in \Cref{sec:architecture} features an LSTM cell of the size $N_{H}=N_{Muntaniala}=96$ and was fabricated in UMC 65 \si{\nano\metre} technology. 
For minimizing leakage, high threshold voltage cells were used.
The quadratic die of $1.57$\si{{\milli\metre}^2} containing a core of $0.93$\si{{\milli\metre}^2}. 
In order to provide enough bandwidth to every LSTM Unit, the parameter memory is split up into $12\times$ SRAM banks (a total of $84 \text{kB}$).
\Cref{fig:shmoo} shows the full operation range from $0.7$\si{\volt} up to $1.275$\si{\volt} achieving a performance of $30.53 \frac{\text{GOP}}{\text{s}}$ in the high-performance operating point and an energy-efficiency of $3.28\frac{\text{TOP/s}}{\text{W}}$ in the high energy-efficiency operating point.
\Cref{tbl:hw_comp} shows the silicon measurement results (e.g., power, maximal frequency) of MUNTANIALA performed on our in-house Advantest SoCV93000 ASIC tester and compares them with other LSTM / RNN ASIC accelerator designs.
Please note that the energy and power results in \Cref{tbl:hw_comp} relate to core power only and excludes the I/O power.

For the technology node of $65\si{\nano\metre}$ \textsc{Muntaniala} is more energy-efficient than most other designs, with three exceptions.
Firstly, Kadetotad \etal~\cite{Kadetotad2020} make use of a \emph{Hierarchical Coarse-Grain Sparsity} (HCGS) weight compression technique by propagating only nonzero weights to the MAC units, allowing them to achieve an energy-efficiency of 8.93 TOP/s/W.
These kinds of compression and other sparsity exploiting approaches are orthogonal to the \textsc{Muntaniala} design and could be combined with the systolic approach in another work.
Yin \etal~\cite{Yin2018} achieve a energy-efficiency of $5.09 \frac{\text{TOP/s}}{\text{W}}$ by using a reconfigurable unit capable of computing CNNs, FCs, and activations for RNNs. 
Their design follows a reloading procedure where the weights are loaded from an off-chip DRAM where the data are stored in a two-symbol Huffman coding compressed fashion. 
For running FC or RNN layers, they state a fully saturated I/O bandwidth.
However, they lack giving any information on the power consumption caused by this high I/O activity.
In contrast, our \textsc{Muntaniala} accelerator has a lower I/O activity than a design based on reloading weights (even in a compressed format).
The 1-16bit weight configurable UNPU~\cite{Lee2019} achieves for 8bits an energy-efficiency of $5.32 \frac{\text{TOP/s}}{\text{W}}$ on the FC layers which is comparable to \textsc{Muntaniala}. 
However, for really computing LSTM networks on their architecture, their on-die SIMD core will most likely become a performance bottleneck and as it computes the non-linear activation functions or element-wise multiplications.

%%%%%%%%%%%%%%%%%%%%%%%%%%%%%%%%%%%%%%%%%%%%
%%%%%%%%%%%%%%%%%%%%%%%%%%%%%%%%%%%%%%%%%%%%

%%%%%%%%%%%%%%%%%%%%%%%%%%%%%%%%%%%%%%%%%%%%
\begin{figure*}[tbp]
\centering
\subfloat[\emph{Slave} (0,0)]{\includegraphics[clip, angle=0, trim= 0pt 0pt 0pt 0pt,  width=0.5\linewidth]{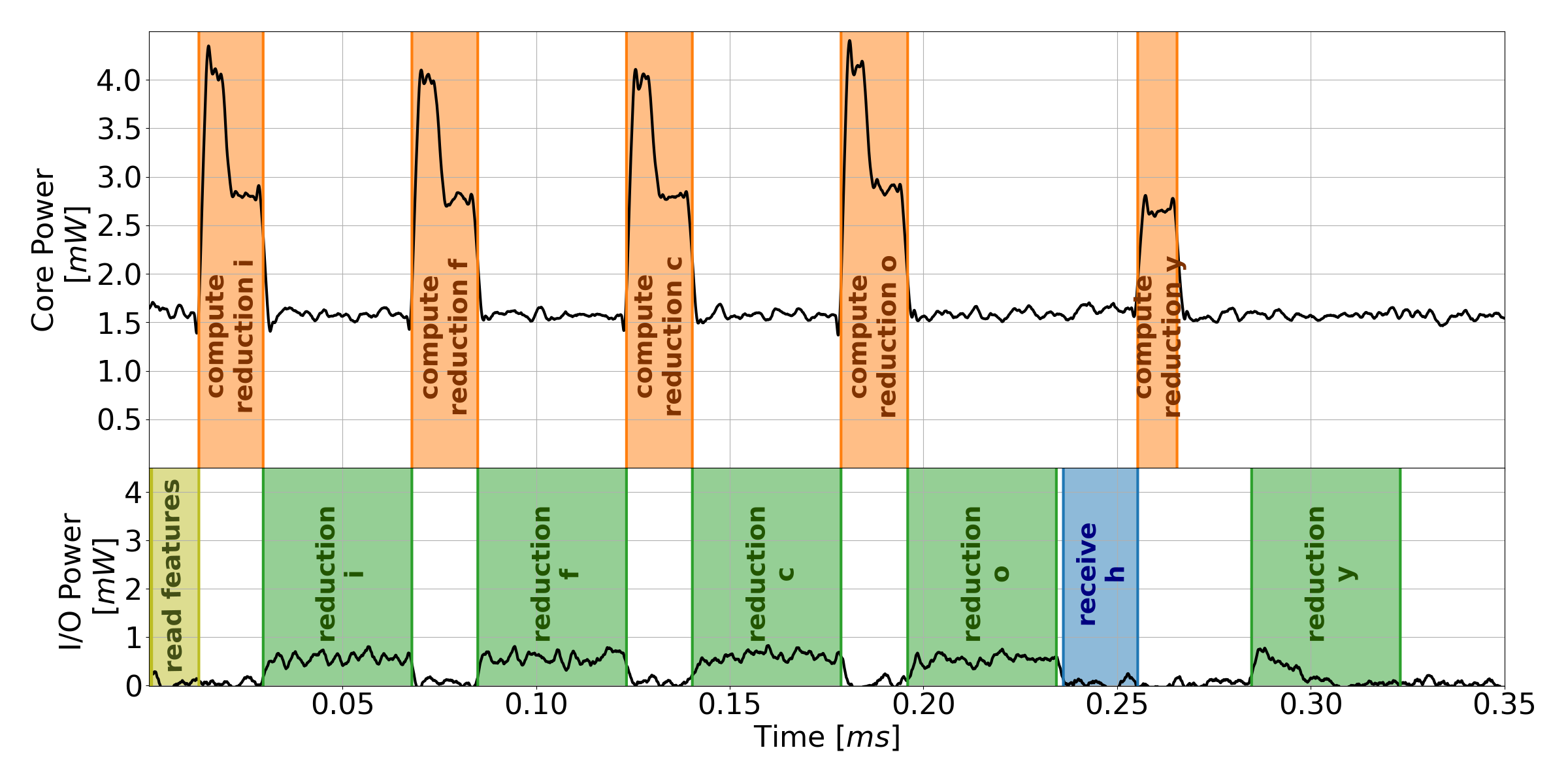}%
\label{fig:power_trace_s00}}
\centering
\subfloat[\emph{Master} (0,1)]{\includegraphics[clip, angle=0, trim= 0pt 0pt 0pt 0pt,  width=0.5\linewidth]{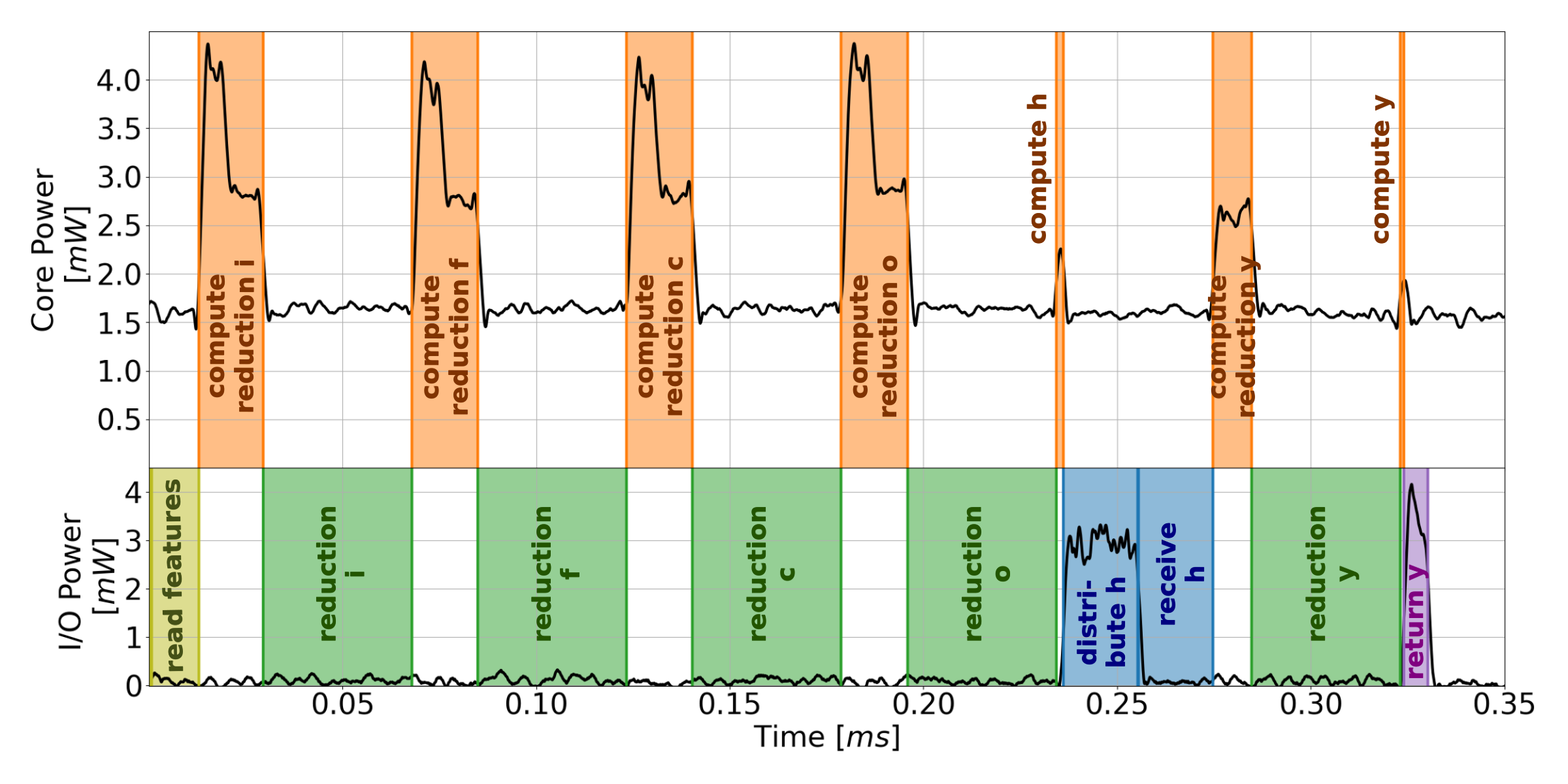}%
\label{fig:power_trace_m01}}
%%%%%
\centering
\caption{Power consumption trace for the core and I/O pads of a \textsc{Muntaniala} accelerator at position (0,0) and at position (0,1) on the \textsc{Vau da Muntanialas} demonstrator running at $10$\,\si{\mega\hertz} with external power supply of $V_{core}=1.2\si{\volt}$, $V_{pad}=2.5\si{\volt}$.
The network is a 2L-192H-123NI where the input features are divided equally on the two columns.}
\label{fig:power_trace}
\end{figure*}
%%%%%%%%%%%%%%%%%%%%%%%%%%%%%%%%%%%%%%%%%%%%

%%%%%%%%%%%%%%%%%%%%%%%%%%%%%%%%%%%%%%%%%%%%
\begin{figure}[tbp]
\centering
\includegraphics[width=0.5\textwidth]{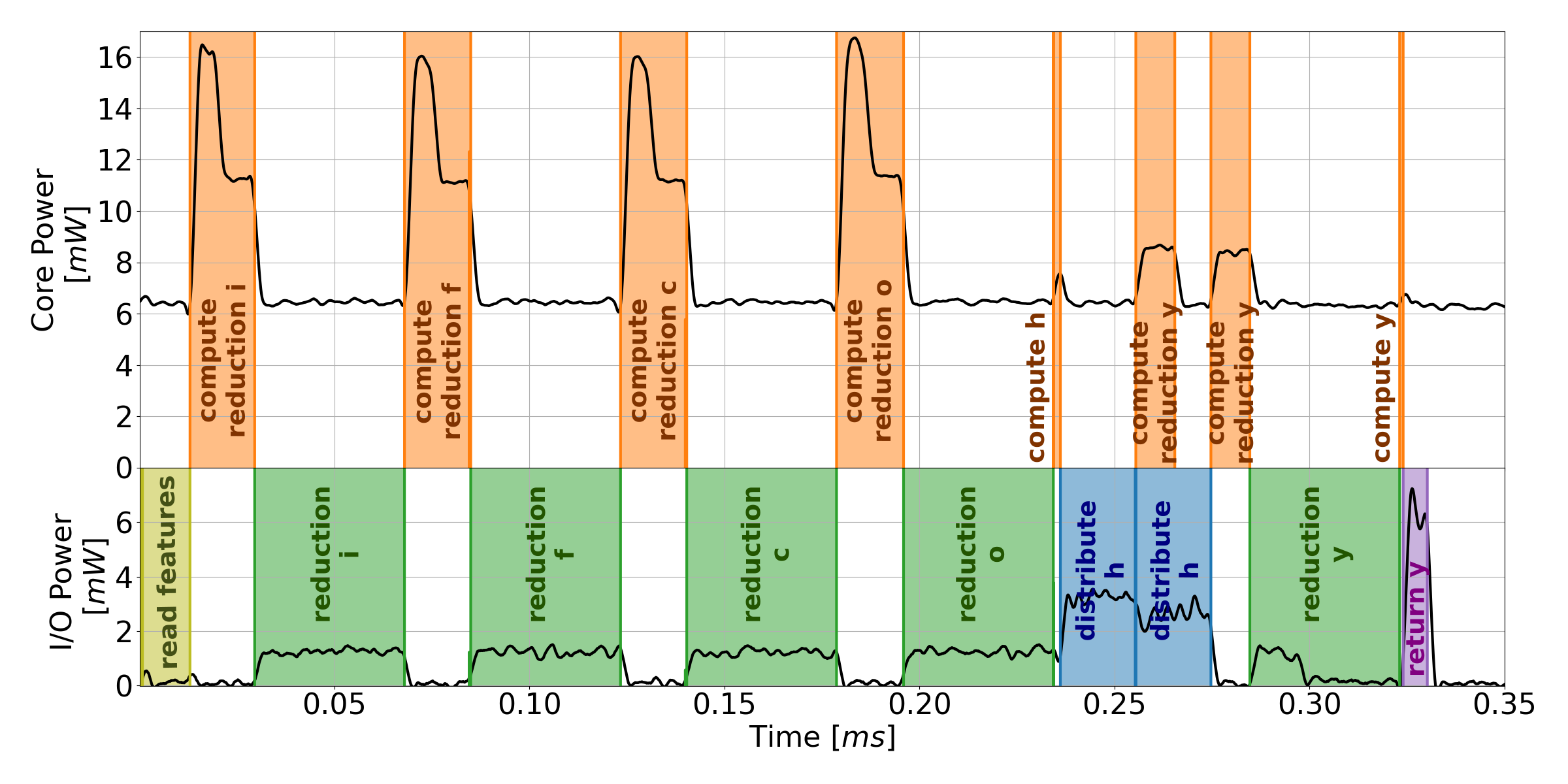}
\caption{Total power consumption for the cores and I/O pads of all four \textsc{Muntaniala} accelerators on the \textsc{Vau da Muntanialas} demonstrator running at $10$\,\si{\mega\hertz} with external power supply of $V_{core}=1.2\si{\volt}$, $V_{pad}=2.5\si{\volt}$.
The network is a 2L-192H-123NI where the input features are divided equally on the two columns.}
\label{fig:powertimeline}
\end{figure}
%%%%%%%%%%%%%%%%%%%%%%%%%%%%%%%%%%%%%%%%%%%%

To compare our design against other designs in a $40\si{\nano\metre}$ technology node, we scaled our $65\si{\nano\metre}$ silicon measurements down to $40\si{\nano\metre}$ technology~\footnote{We followed a simple technology-based feature size $l_{i}$ and standard supply voltage $V_{dd,i}$ scaling:  $P_{new}=P_{old}\cdot(l_{new}/l_{old})(V_{dd,new}/V_{dd,old})^2$.}.
Even with a very conservative scaling where the achieved frequency is kept the same as for the measured frequency in the low-power operating point in $65\si{\nano\metre}$ \textsc{Muntaniala} is approximately 1.4$\times$ more energy-efficient than the design proposed by Wu \etal~\cite{Wu2019}.

On another aspect, these results are only applicable to tasks requiring reasonably small LSTM networks.
Most of the various accelerator designs listed in \Cref{tbl:hw_comp} have an on-chip SRAM memory size between 10-348kB.
Once the task complexity and the corresponding network size increases, these accelerators are forced to reload all parameters, which cripples the throughput or results in heavily I/O-boundness and significantly degrades their energy-efficiency.
In this scenario, the systolic design of \textsc{Muntaniala} shows its real advantages.

Even though some other works achieve comparable energy-efficiency, \textsc{Muntaniala} and \textsc{Chipmunk} are more area efficient than all accelerator designs in the same technology node of $65\si{\nano\metre}$, and only Wu \etal~\cite{Wu2019} design in a more dense technology of $40\si{\nano\metre}$ is achieving a higher area efficiency.

FPGA accelerators such as~\cite{Gao2018} or ESE~\cite{Han2017} exploit algorithmic compression and sparsity schemes. Nevertheless, they achieve 1-2 orders of magnitude lower energy efficiency than the ASIC implementations shown in \Cref{tbl:hw_comp}: 165\,GOP/s/W~\cite{Gao2018} and 120\,GOP/s/W~\cite{Han2017} (results for the corresponding dense LSTM network).

To summarize, our \textsc{Muntaniala} is, in its stand-alone performance, the most area-efficient LSTM accelerator. Our architecture could benefit from efficiency improvements as achieved by accelerators exploiting algorithmic advancements such as the HCGS compression technique used by the most energy-efficient accelerator~\cite{Kadetotad2020}.  However our main focus and unique contribution is not on core efficiency, but on  providing a solution of beyond-die-scaling to tackle larger networks.

%%%%%%%%%%%%%%%%%%%%%%%%%%%%%%%%%%%%%%%%%%%%%%%%%%%%%%%%%%%%%%%%%%%%%%%%%%%%%%%%%%%%%%%%
\subsection{\textsc{Vau da Muntanialas} - Systolic Array} \label{subsec:vau}

\Cref{fig:power_trace,fig:powertimeline} show the power consumption during all phases of computation, as explained in \Cref{subsubsec:arch_overview} and shown in \Cref{fig:timeline}.
Measurements were taken with a high-sensitivity current probe and a Keysight Oscilloscope.
Due to size constraints, we use only a 1L-192NH-123NI subset of the fully trained 3L-384NH-123NI network.
For the inference, preprocessed input features for samples from the TIMIT dataset were used.
\textsc{Muntaniala} has separate I/O and core supply pads, each of which have dedicated test points on the demonstrator PCB \textsc{Vau da Muntanialas}.
For every power measurement, we make use of these test points and measure I/O and core power for each chip individually (\Cref{fig:power_trace}), or for all chips at once (for \Cref{fig:powertimeline})
The various computation and communication phases are highlighted according to the measured active handshakes.

\Cref{fig:power_trace} shows the power traces for a \textsc{Muntaniala} accelerator at position (0,0) and (0,1), furthermore called \emph{slave} S(0,0) and \emph{master} M(0,1).
In the first phase, every chip receives the new features by the FPGA.
During the reduction phases (green), the output I/O pads of S(0,0) are driving the reduction input I/O pads of M(0,1) (see \Cref{fig:pcb_connection,subsubsec:arch_overview}).
The reduction phases are repeated four times for each gate $\mathbf{i}_{t}$, $\mathbf{f}_{t}$, $\mathbf{o}_{t}$ and the cell candidate state $\mathbf{\tilde{c}}_{t}$.
After the hidden state $\mathbf{h}_{t}$ computation, the chip at position M(1,1) distributes its partial state to M(0,1). 
After that M(0,1) distributes its own partial state to S(0,0) and S(1,0).
These two hidden state distribution phases are highlighted in blue.
After a last reduction round for the FCL, M(0,1) and M(1,1) write out their result back to the FPGA.

In general, the measurements reveal distinct power traces for computation and communication phases, respectively. Notably, the computation phases consume more energy than the I/O communication. 
In reduction phases (green), S(0,0) outputs partial results that are read by M(0,1), which is shown in our measurements by a power level difference.
We measured the energy consumption of the I/O supply of S(0,0) and M(0,1) separately for the duration of the communication phases and report an driving costs of 27.8\,$\frac{\si{\pico\joule}}{\text{bit}}$ and a receiving cost of 4.7 $\frac{\si{\pico\joule}}{\text{bit}}$.
This behavior stems from the typically large inverters within the output pad drivers capable of driving high currents for the off-chip wiring.
On the other hand, input pads still consume some energy as they have overvoltage protection and some smaller inverters to convert the incoming voltage level of $2.5\si{\volt}$ to the voltage level $1.2\si{\volt}$ of the CMOS cells in the core.
We note that the measurements of the I/O pad energy are consistent with other published measurements performed on PCBs~\cite{Okuhara2020}.

%%%%%%%%%%%%%%%%%%%%%%%%%%%%%%%%%%%%%%%%%%%%
%%%%%%%%%%%%%%%%%%%%%%%%%%%%%%%%%%%%%%%%%%%%
\begin{table*}[t]
\caption{
Extrapolated inference time, power, and energy consumption for various systolic grid and LSTM network sizes. The estimations rely on measurements performed at 10MHz with Vcore=1.2V, Vpad=2.5V.
}
\label{tbl:energy_eval}
\centering
\begin{tabular}{@{}ccccrrrrrr@{}}
\toprule
\multicolumn{2}{c}{\textbf{Network Size}} & \multicolumn{2}{c}{\textbf{\# Chips}} & \multicolumn{1}{c}{\textbf{Time per}} & \multicolumn{1}{c}{\textbf{Power}} & \multicolumn{3}{c}{\textbf{Inference Energy}} & \multicolumn{1}{c}{\textbf{\% I/O}} \\
\multicolumn{1}{c}{\textbf{\#Layer}} & \multicolumn{1}{c}{\textbf{Hidden State}} & \multicolumn{1}{c}{\textbf{per Layer}} & \multicolumn{1}{c}{\textbf{Total}} & \multicolumn{1}{c}{\textbf{inference {[}\si{\micro\second}{]}}} & \multicolumn{1}{c}{\textbf{Cores {[}\si{\milli\watt}{]}}} & \multicolumn{1}{c}{\textbf{Cores {[}\si{\micro\joule}{]}}} & \multicolumn{1}{c}{\textbf{I/O {[}\si{\micro\joule}{]}}} & \multicolumn{1}{c}{\textbf{Total {[}\si{\micro\joule}{]}}} & \multicolumn{1}{c}{\textbf{Contribution}} \\
\midrule
1 & 96  & 1x1 & 1  & 101.2  & 2.0   & 0.2   & 0.0  & 0.2   & 5.9  \\
1 & 56  & 1x1 & 1  & 81.2   & 2.0   & 0.2   & 0.0  & 0.2   & 6.1  \\
\midrule
1 & 192 & 2x2 & 4  & 295.2  & 7.9   & 2.3   & 0.3  & 2.6   & 12.1 \\
1 & 288 & 3x3 & 9  & 469.8  & 17.7  & 8.3   & 1.0  & 9.3   & 10.4 \\
1 & 384 & 4x4 & 16 & 644.4  & 31.5  & 20.3  & 2.1  & 22.3  & 9.2  \\
1 & 480 & 5x5 & 25 & 819.0  & 49.2  & 40.3  & 3.7  & 43.9  & 8.3  \\
\midrule
2 & 96  & 1x1 & 2  & 182.8  & 3.9   & 0.7   & 0.1  & 0.8   & 7.3  \\
2 & 192 & 2x2 & 8  & 532.0  & 15.7  & 8.4   & 0.8  & 9.2   & 8.6  \\
\midrule
3 & 384 & 4x4 & 48 & 1'933.2 & 94.4  & 182.6 & 11.2 & 193.8 & 5.8  \\
3 & 480 & 5x5 & 75 & 2'457.0 & 147.6 & 362.6 & 21.0 & 383.5 & 5.5 \\
\bottomrule
\end{tabular}
\end{table*}
%%%%%%%%%%%%%%%%%%%%%%%%%%%%%%%%%%%%%%%%%%%%
%%%%%%%%%%%%%%%%%%%%%%%%%%%%%%%%%%%%%%%%%%%%

\Cref{fig:powertimeline} shows the total power trace of all four \textsc{Muntanialas}.
On average, we measure a total core power of 7.87 mW and total I/O power of 1.13 \si{\milli\watt} over a time of 330$\mu$s, given a total average system power of 9.00 \si{\milli\watt} for the \textsc{Vau da Muntanialas} FMC card computing a network of the size 1L-192NH-123NI at 10MHz, 1.2V core supply, and 2.5V pad supply.
An inference on \textsc{Vau da Muntanialas} consumes 2.97\si{\micro\joule} with the output $\mathbf{y}_{t}$ computation and 2.74\si{\micro\joule} without the output $\mathbf{y}_{t}$ computation.
% Including the output $\mathbf{y}_{t}$ computation, an inference on \textsc{Vau da Muntanialas} consumes 2.97\si{\micro\joule} and 2.74\si{\micro\joule} without the output $\mathbf{y}_{t}$ computation.
11.9\% and 12.5\% respectively are consumed by the I/O, and 87.5\% and 88.1\% by the cores.

%%%%%%%%%%%%%%%%%%%%%%%%%%%%%%%%%%%%%%%%%%%%%%%%%%%%%%%%%%%%%%%%%%%%%%%%%%%%%%%%%%%%%%%%
\subsection{Systolic Array - System Energy Evaluation} \label{subsec:systempower}
For applications that are more demanding than keyword spotting, such as phoneme recognition, a bigger grid than $2\times 2$ \textsc{Muntaniala} dies would be necessary.
With the core and I/O power consumption on the \textsc{Vau da Muntanialas} demonstrator measured in \Cref{subsec:vau}, we estimate the overall power consumption for various systolic grid sizes in \Cref{tbl:energy_eval}.
For simplicity reasons, we chose $N_I=N_H$ and skipped the configuration phase and the output computation, which depends on $N_O$.
We stress that master dies are connected to slaves, other masters, and the FPGA at the same time. This leads to extreme wire lengths compared to normal slave-master wires.
To keep our numbers comparable, we removed the energy consumption caused by the additional load of the longer wires to the FPGA.
In a real-world scenario, the FPGA could be relocated on the same board and thus have vastly shorter wiring.
The inference time is measured and scaled for each operation phase from RTL simulation measurements performed for a systolic grid of 1x1, 2x2 \textsc{Muntaniala} chips. The core power is extrapolated from the measurements performed on the demonstrator \textsc{Vau da Muntanialas} (see \Cref{subsec:vau}). The I/O power estimation considers the measured energy cost for driving and receiving data and leakage (see \Cref{subsec:vau}) and considers toggling statistics for each interface based on a layer of the trained network.

Whenever the LSTM network fits completely onto one single die, the power is almost entirely dominated by the core (94.1\% and 93.9\%) because no communication for the reduction and hidden state redistribution is needed.
On a systolic grid, these phases contribute to the I/O power.
However, they consume less energy than naively reloading all parameters.
Note that smaller systolic arrays lead to a higher contribution of the I/O to the total energy consumption, in the worst case up to 12.1\%.
Running inference on a network of the size 3L-384NH-123NI (similar to Graves~\etal~\cite{AlexGraves2013}) requires a grid of 48 \textsc{Muntaniala} dies.
The inference takes 1.9\si{\milli\second} consuming 193.8$\si{\pico\joule}$ at a frequency of 10\si{\mega\hertz}.
If more performance is required, the \textsc{Muntaniala} dies could run with up to 159\si{\mega\hertz}  leading to $122$\si{\micro\second} per inference.
Note that the \textsc{Muntaniala} prototype was designed as a proof of concept.
For industrial implementation, the die size, and with it the hidden state size on a single \textsc{Muntaniala} die, could be scaled up further before scaling multi-die.

%%%%%%%%%%%%%%%%%%%%%%%%%%%%%%%%%%%%%%%%%%%%%%%%%%%%%%%%%%%%%%%%%%%%%%%%%%%%%%%%%%%%%%%%
%%%%%%%%%%%%%%%%%%%%%%%%%%%%%%%%%%%%%%%%%%%%%%%%%%%%%%%%%%%%%%%%%%%%%%%%%%%%%%%%%%%%%%%%
\section{Conclusion}\label{sec:conclusion}
We have presented \textsc{Muntaniala}, a systolically scalable hardware architecture for various types of LSTM neural networks, dramatically reducing the need for parameter reloading and therefore drastically minimizes I/O energy consumption.
Additionally, we presented \textsc{Vau da Muntanialas}, to the best of our knowledge the first complete hardware demonstration of a multi-chip array for RNNs using our \textsc{Muntaniala} LSTM accelerator. Specifically, we combined four identical \textsc{Muntaniala} accelerators in a grid of 2$\times$2 chips, performing LSTM inference with 192 hidden states in 330 \si{\micro\second} with a core power of 7.87 \si{\milli\watt} and a I/O power of 1.13 \si{\milli\watt} consuming 2.95 \si{\micro\joule} at 10 \si{\mega\hertz}.
Our evaluation shows that running an inference on a 3L-384NH-123NI network (similar to Graves \etal~\cite{AlexGraves2013}), a grid of 48 \textsc{Muntaniala} dies is needed which can perform inference in $122$\si{\micro\second} at a frequency of 159\si{\mega\hertz}, and 1.9\si{\milli\second} consuming 193.8$\si{\pico\joule}$ at a frequency of 10\si{\mega\hertz}.
We estimate the I/O contribution around 5.8\%.

%%%%%%%%%%%%%%%%%%%%%%%%%%%%%%%%%%%%%%%%%%%%%%%%%%%%%%%%%%%%%%%%%%%%%%%%%%%%%%%%%%%%%%%%
%%%%%%%%%%%%%%%%%%%%%%%%%%%%%%%%%%%%%%%%%%%%%%%%%%%%%%%%%%%%%%%%%%%%%%%%%%%%%%%%%%%%%%%%
\section*{Acknowledgment}
We would like to thank Pascal Alexander Hager for his support on the PCB design and FPGA controller implementation.
Additionally, we thank Justin MacPherson for his contribution of the training part, and Igor Susmelj for his contributions to the design and implementation of the predecessor \textsc{Chipmunk}.
This work has been supported in part by “Heterogenous Computing Systems with Customized Accelerators” (IZHRZ0\_180625) project supported by the Swiss National Science Foundation as part of the Croatian-Swiss Research Program.

%%%%%%%%%%%%%%%%%%%%%%%%%%%%%%%%%%%%%%%%%%%%%%%%%%%%%%%%%%%%%%%%%%%%%%%%%%%%%%%%%%%%%%%%
%%%%%%%%%%%%%%%%%%%%%%%%%%%%%%%%%%%%%%%%%%%%%%%%%%%%%%%%%%%%%%%%%%%%%%%%%%%%%%%%%%%%%%%%
% references section
\bibliographystyle{IEEEtran}
\bibliography{main.bbl}

\newpage
%%%%%%%%%%%%%%%%%%%%%%%%%%%%%%%%%%%%%%%%%%%%%%%%%%%%%%%%%%%%%%%%%%%%%%%%%%%%%%%%%%%%%%%%
%%%%%%%%%%%%%%%%%%%%%%%%%%%%%%%%%%%%%%%%%%%%%%%%%%%%%%%%%%%%%%%%%%%%%%%%%%%%%%%%%%%%%%%%
\begin{IEEEbiography}[{\includegraphics[width=1in,height=1.25in,clip,keepaspectratio]{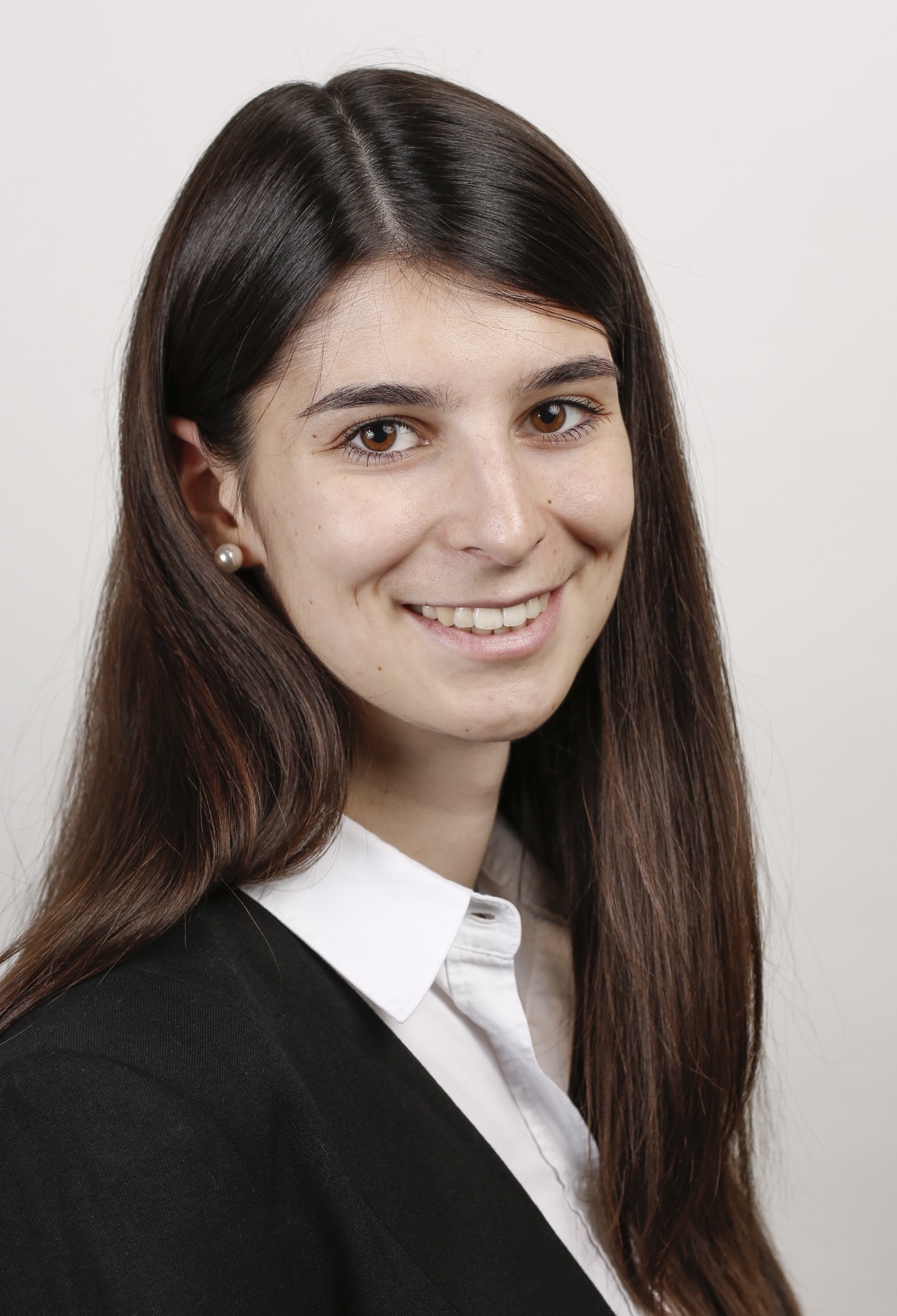}}]{Gianna Paulin}
received her B.Sc. and M.Sc. degrees in in Electrical Engineering and Information Technology from the Swiss Federal Institute of Technology Zurich (ETH Zürich), Switzerland, where she started as a Ph.D. student with the Integrated Systems Laboratory at the beginning of 2019. Her main interests lay in reduced precision deep learning from the algorithmic and hardware acceleration aspect with a focus on time series applications and low power embedded systems.
\end{IEEEbiography}

\begin{IEEEbiography}[{\includegraphics[width=1in,height=1.15in,clip,keepaspectratio]{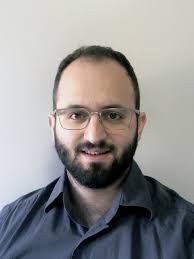}}]{Francesco Conti} received the Ph.D. degree in electronic engineering from the University of Bologna, Italy, in 2016. He is currently an Assistant Professor in the DEI Department of the University of Bologna. From 2016 to 2020, he held a research grant in the DEI department of University of Bologna and a position as postdoctoral researcher at the Integrated Systems Laboratory of ETH Z\"urich in the Digital Systems group.
His research focuses on the development of advanced deep learning based intelligence on top of ultra-low power, ultra-energy efficient programmable Systems-on-Chip -- from both the hardware and software perspective.
His work has resulted in more than 40 publications in international conferences and journals and has been awarded several times, including the 2020 IEEE TCAS-I Darlington Best Paper Award.
\end{IEEEbiography}

% if you will not have a photo at all:
\begin{IEEEbiography}[{\includegraphics[width=1in,height=1.25in,clip,keepaspectratio]{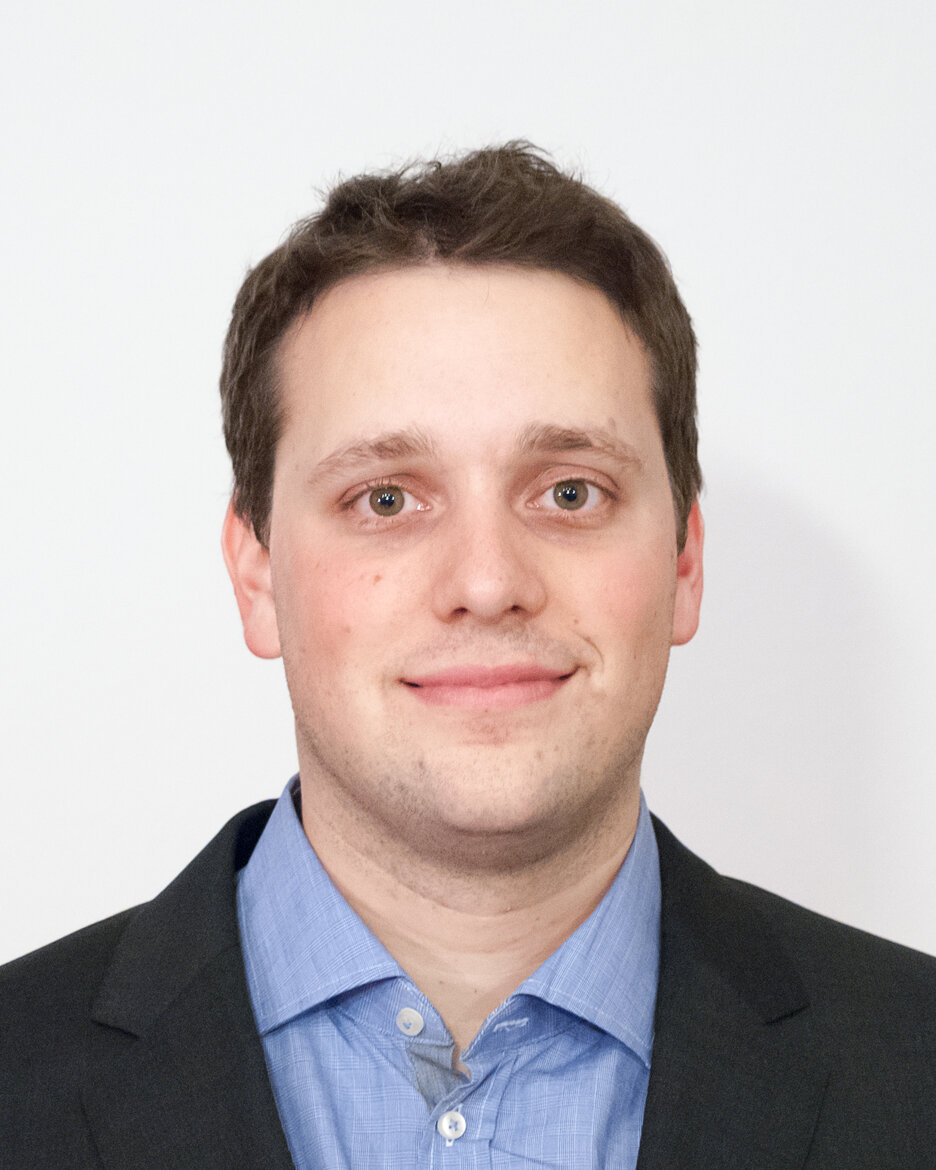}}]{Lukas Cavigelli}
received the B.Sc., M.Sc., and Ph.D. degree in electrical engineering and information technology from ETH Zürich, Zürich, Switzerland in 2012, 2014 and 2019, respectively. After spending a year as a Postdoc at ETH Zürich, he has joined Huawei's Zurich Research Center in Spring 2020. His research interests include deep learning, computer vision, embedded systems, and low-power integrated circuit design. He has received the best paper award at the VLSI-SoC and the ICDSC conferences in 2013 and 2017, the best student paper award at the Security+Defense conference in 2016, and the Donald O. Pederson best paper award (IEEE TCAD) in 2019.
\end{IEEEbiography}

\begin{IEEEbiography}[{\includegraphics[width=1in,height=1.25in,clip,keepaspectratio]{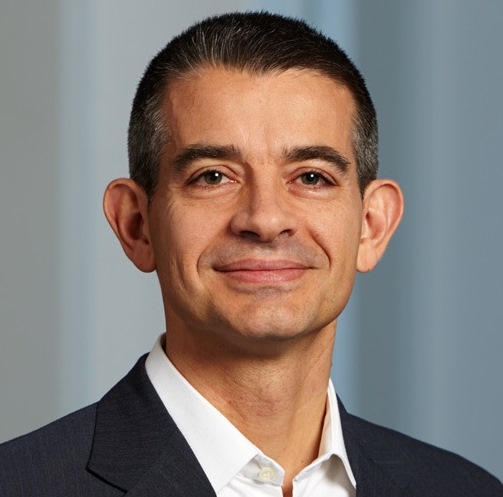}}]{Luca Benini}
is the Chair of Digital Circuits and Systems at ETH Zürich and a Full Professor at the University of Bologna. He has served as Chief Architect for the Platform2012 in STMicroelectronics, Grenoble. Dr. Benini’s research interests are in energy-efficient system and multi-core SoC design. He is also active in the area of energy-efficient smart sensors and sensor networks. He has published more than 1’000 papers in peer-reviewed international journals and conferences, four books and several book chapters. He is a Fellow of the ACM and of the IEEE and a member of the Academia Europaea.
\end{IEEEbiography}

\vfill

\end{document}